\documentclass{article}
\usepackage[T1]{fontenc}

\usepackage[margin=1in]{geometry}
\usepackage{libertine} 
\usepackage{amsmath,amssymb,amsthm}
\usepackage{graphicx}
\usepackage{xcolor}

\usepackage{hyperref}
\definecolor{mydarkblue}{rgb}{0,0.08,0.85}
\hypersetup{colorlinks=true,
linkcolor=mydarkblue,
citecolor=mydarkblue,
filecolor=mydarkblue,
urlcolor=mydarkblue,
bookmarksopen=true}

\usepackage{pythonhighlight}
\usepackage{wrapfig}
\usepackage{xspace}
\usepackage{subcaption}
\usepackage{comment}
\usepackage{todonotes}
\usepackage{relsize}
\usepackage{nicefrac}
\usepackage[T1]{fontenc}
\usepackage{booktabs}
\usepackage{float}

\usepackage[normalem]{ulem}
\usepackage{xcolor}

\newcommand{\scrcmd}{\textsuperscript}
\newcommand{\scr}[1]{\scrcmd{#1}}

\usepackage[backend=biber,style=alphabetic,natbib=true,minalphanames=3]{biblatex}

\newcommand{\sandbox}{\textit{3DB}}

\title{\sandbox{}: A Framework for Debugging Computer Vision Models}
\newcommand\AND{
    \end{tabular}\hfil\linebreak[4]\hfill%
    \begin{tabular}[t]{c}\ignorespaces%
}
\author{
    Guillaume Leclerc\footnotemark[2] \\
    {\normalsize \textsc{leclerc@mit.edu}} \\
    MIT \footnote{Work partially completed while at Microsoft Research.}
    \and
    Hadi Salman\footnote{Equal contribution.} \\
    {\normalsize \textsc{hady@mit.edu}} \\
    MIT \footnotemark[1]
    \and
    Andrew Ilyas\footnotemark[2] \\
    {\normalsize \textsc{ailyas@mit.edu}} \\
    MIT \\
    \AND
    Sai Vemprala \\
    {\normalsize \textsc{saihv@microsoft.com}} \\
    Microsoft Research
    \and 
    Logan Engstrom \\
    {\normalsize \textsc{engstrom@mit.edu}} \\
    MIT
    \and
    Vibhav Vineet \\
    {\normalsize \textsc{vivineet@microsoft.com}} \\
    Microsoft Research
    \and 
    Kai Xiao \\
    {\normalsize \textsc{kaix@mit.edu}} \\
    MIT
    \and
    Pengchuan Zhang \\
    {\normalsize \textsc{penzhan@microsoft.com}} \\
    Microsoft Research
    \and 
    Shibani Santurkar \\
    {\normalsize \textsc{shibani@mit.edu}} \\
    MIT
    \and
    Greg Yang \\
    {\normalsize \textsc{ge.yang@microsoft.com}} \\
    Microsoft Research
    \and
    Ashish Kapoor \\
    {\normalsize \textsc{akapoor@microsoft.com}} \\
    Microsoft Research
    \and 
    Aleksander M\k{a}dry \\
    {\normalsize \textsc{madry@mit.edu}} \\
    MIT 
}
\date{}

\begin{document}
    \maketitle
    \begin{abstract}
    \normalsize
We introduce \sandbox: an extendable, unified
framework for testing and debugging vision models using photorealistic simulation. 
We demonstrate, through a wide range 
of use cases, that \sandbox{} allows users to discover vulnerabilities in 
computer vision systems and gain insights into how models make decisions.
\sandbox{} captures and generalizes many robustness analyses from prior work, 
and enables one to study their interplay. Finally, we find that the 
insights
generated by the system transfer to the physical world.

We are releasing \sandbox{} as a
library\footnote{\url{https://github.com/3db/3db}} alongside a set of example
analyses\footnote{\url{https://github.com/3db/blog_demo}},
guides\footnote{\url{https://3db.github.io/3db/usage/quickstart.html}}, and
documentation\footnote{\url{https://3db.github.io/3db/}}.
    \end{abstract}

    \clearpage
    \section{Introduction}
    \label{sec:intro}
    Modern machine learning models turn out to be remarkably
brittle under distribution shift. Indeed, in the context of computer
vision, models exhibit an abnormal sensitivity to slight input rotations and
translations \citep{engstrom2019rotation,kanbak2018geometric}, synthetic image corruptions
\citep{hendrycks2019benchmarking,kang2019testing}, and changes to the
data collection pipeline \citep{recht2018imagenet, engstrom2020identifying}. 
Still, while such brittleness is widespread, it is often hard to understand its
root causes, or even to characterize the precise situations in which this
unintended behavior arises. 

\begin{figure}[!pb]
        \centering
        \includegraphics[trim={4cm 15cm 4cm 0},clip,page=1,width=\textwidth]{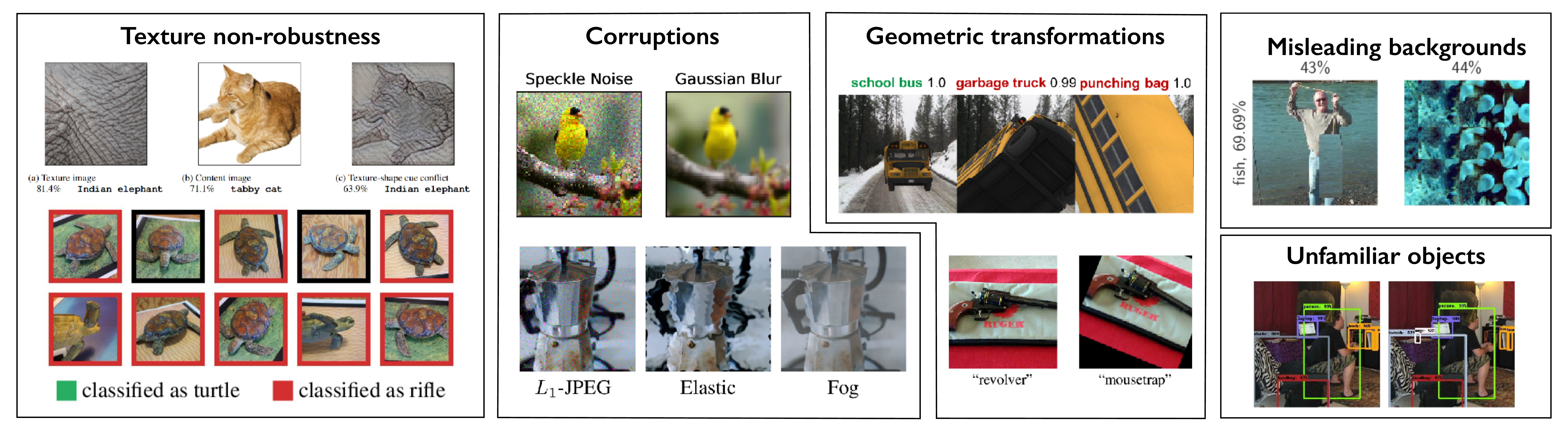}
        \caption{Examples of vulnerabilities of computer vision systems
        identified through prior in-depth robustness studies. Figures reproduced from
        \citep{geirhos2018imagenettrained,athalye2018synthesizing,
        hendrycks2019benchmarking,kang2019testing,
        alcorn2019strike,engstrom2019rotation,xiao2020noise,rosenfeld2018elephant}.}
        \label{fig:vis_samples}
    \end{figure}

How do we then comprehensively diagnose model failure modes? 
Stakes are often too high to simply deploy models and collect eventual
``real-world'' failure cases. 
There has thus been a line of work in computer vision
focused on identifying systematic sources of model failure such as
unfamiliar object orientations \citep{alcorn2019strike}, misleading
backgrounds \citep{zhu2017object, xiao2020noise}, or shape-texture conflicts
\citep{geirhos2018imagenettrained, athalye2018synthesizing}. 
These analyses---a selection of which is 
visualized in Figure \ref{fig:vis_samples}---reveal patterns or
situations that degrade performance of vision models, providing invaluable 
insights into model robustness. 
Still, carrying out each such analysis
requires its own set of (often complex) tools and techniques, usually
accompanied by a significant amount of manual labor (e.g., image editing, style
transfer, etc.), expertise, and data cleaning. 
This prompts the question:

\begin{center}
{\em Can we support reliable discovery of
model failures in a systematic, automated, and unified way?}
\end{center}

\begin{figure}[!pb]
        \centering
        \includegraphics[trim={3cm 25cm 3cm 0},clip,page=3,width=\textwidth]{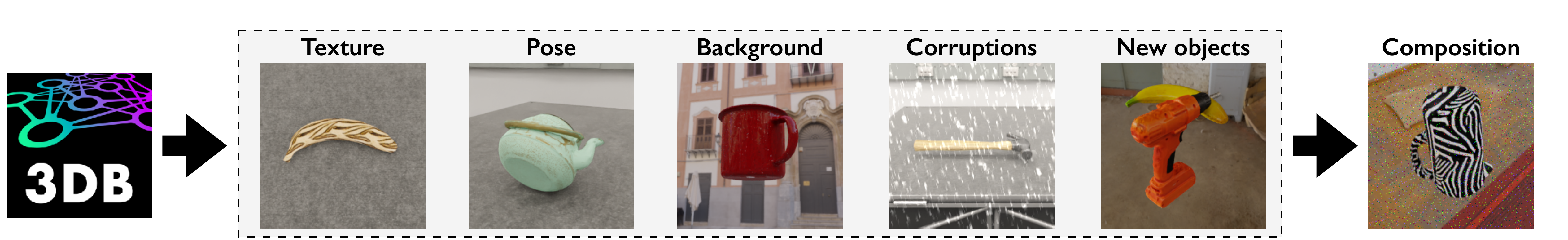}
        \caption{The \sandbox{} framework is modular enough to facilitate---among
        other tasks---efficient rediscovery of all the types of
        brittleness shown in Figure \ref{fig:vis_samples} in an integrated
        manner. It also allows users to realistically compose transformations
        (right) while still being able to disentangle the results.} 
        \label{fig:reproduced}
    \end{figure}

\paragraph{Contributions.} In this work, we propose \sandbox, a framework for
automatically identifying and analyzing the failure modes of computer vision
models. This framework makes use of a 3D simulator to render realistic
scenes that can be fed into any computer vision system. Users can specify a 
set of transformations to apply to the scene---such as pose changes, 
background changes, or camera effects---and can also customize and 
compose them. The system then performs a guided search, 
evaluation, and aggregation over these user-specified configurations and 
presents the user with an
interactive, user-friendly summary of the model's performance and
vulnerabilities. \sandbox{} is general enough to enable users to, with
little-to-no effort, re-discover insights from prior work on robustness to
pose, background, and texture bias (cf. Figure~\ref{fig:reproduced}), among 
others. 
Further, while prior studies have largely been focused on examining model 
sensitivities along a single axis, \sandbox{} allows users to compose 
various transformations to understand the interplay between them, while still 
being able to disentangle their individual effects. 


The remainder of this paper is structured into the following three parts: in 
Section
\ref{sec:design} we discuss the design of \sandbox, including the motivating
principles, design goals, and concrete architecture used. We highlight how the
implementation of \sandbox{} allows users to quickly experiment, stress-test, and
analyze their vision models. Then, in Section \ref{sec:evaluation} we illustrate
the utility of \sandbox{} through a series of case studies uncovering biases in
an ImageNet-pretrained classifier.
Finally, we show (in Section \ref{sec:realworld-exp}) that the vulnerabilities
uncovered with \sandbox{} correspond to actual failure modes in the physical
world (i.e., they are not specific to simulation).

    \section{Designing \sandbox}
    \label{sec:design}
    The goal of \sandbox{} is to leverage photorealistic simulation in order to
effectively diagnose failure modes of computer vision
models.  
To this end, the following set of principles guide the design of \sandbox{}:
\begin{itemize}
    \item[(a)] {\bf Generality}: \sandbox{} should support any type of computer vision
    model (i.e., not necessarily a neural network) trained on any dataset and 
    task (i.e., not necessarily classification). Furthermore, the
    framework should support diagnosing non-robustness
    with respect to any parameterizable three-dimensional scene transformation. 
    \item[(b)] {\bf Compositionality}: Data corruptions
    and transformations
    rarely occur in isolation. 
    Thus, \sandbox{} should allow users to investigate robustness along many different
    axes simultaneously.
    \item[(c)] {\bf Physical realism}: The vulnerabilities
    extracted from \sandbox{} should correspond to models' behavior in the real 
    (physical) world, and, in particular, not depend on artifacts of the simulation process itself.
    Specifically, the insights that \sandbox{} produces should not be affected by a simulation-to-reality
    gap, and still hold when models are deployed in the wild.
    \item[(d)] {\bf User-friendliness}: \sandbox{} should be simple to use and
    should relay insights to the user
    in an easy-to-understand manner. 
    Even non-experts should be
    able to look at the result of a \sandbox{} experiment and easily understand
    what the weak points of their model are, as well as gain insight into
    how the model behaves more generally.
    \item[(e)] {\bf Scalability}:
    \sandbox{} should be performant and parallelizable.
\end{itemize}

\subsection{Capabilities and workflow}
\label{sec:workflow}

\begin{figure}[!h]
    \centering
    \includegraphics[trim={1cm 0 0 1cm},clip,page=5,width=\textwidth]{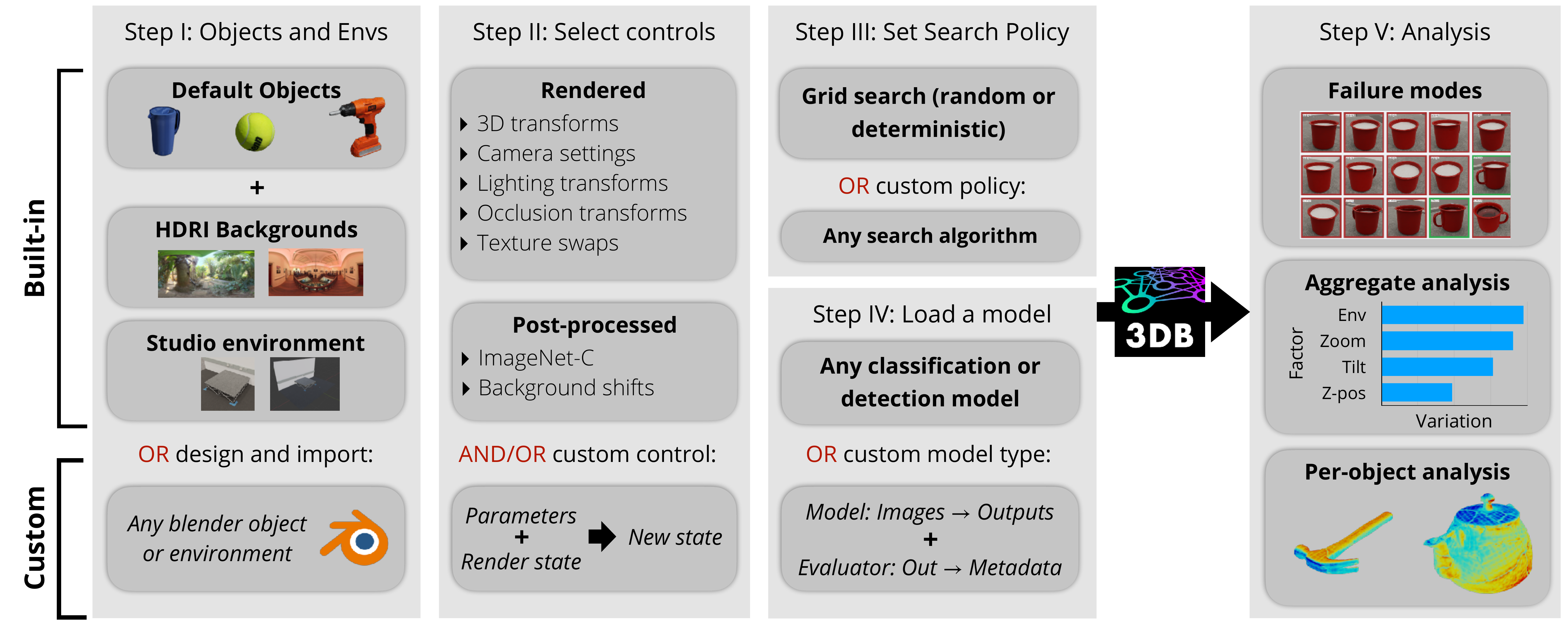}
    \caption{An overview of the \sandbox{} workflow: First, the user specifies 
    a set of 3D object models and environments to use for debugging. The 
    user also enumerates a set of (in-built or custom) transformations, known 
    as controls, to be applied by \sandbox{} while rendering the scene. Based 
    on a user-specified search policy over all these controls (and their 
    compositions),  \sandbox{} then selects the exact scenes 
    to render. The computer vision model is finally evaluated on these scenes 
    and the results are logged in a 
    user-friendly manner in a custom dashboard.} 
    \label{fig:workflow}
\end{figure}

To achieve the goals articulated above, we design \sandbox{} in a modular
manner, i.e., as a combination of swappable components. This combination allows
the user to specify transformations they want 
to test, search over the space of these transformations, and aggregate the 
results of this search in a concise way. More specifically, 
the \sandbox{} workflow revolves around five steps (visualized in Figure \ref{fig:workflow}):
\begin{enumerate} 
    \item {\bf Setup}: The user collects one or more 3D meshes that correspond
    to objects the model is trained to recognize, as well as a set of
    environments to test against. 
    \item {\bf Search space design}: The user defines a {\em search
    space} by specifying a set of transformations (which \sandbox{} calls
    {\em controls}) that they expect the computer vision model to be robust
    to (e.g., rotations, translations, zoom, etc.). Controls are grouped into
    ``rendered controls'' (applied during the rendering process) and
    ``post-processor controls'' (applied after the rendering as a 2D image
    transformation). 
    \item {\bf Policy-guided search}: After the user has specified a set of
    controls, \sandbox{} instantiates and renders a myriad of object
    configurations derived from compositions of the given transformations. It
    records the behavior of the ML model on each constructed scene for later
    analysis. 
    A user-specified {\em search policy} over the space of all possible
    combinations of transformations determines the exact scenes for
    \sandbox{} to render.
    \item {\bf Model loading}: The only remaining step before running a
    \sandbox{} analysis is loading the vision model that the user wants to
    analyze (e.g., a pre-trained classifier or object detection model).  
    \item {\bf Analysis and insight extraction}: 
    Finally, \sandbox{} is equipped with a model
    {\em dashboard} (cf. Appendix \ref{app:dashboard}) that can read the
    generated log files and produce a user-friendly visualization of the
    generated insights.  
    By default, the dashboard has three panels. The first of these is failure mode
    display, which highlights configurations, scenes, and transformations that
    caused the model to misbehave. 
    The per-object analysis pane allows the user to inspect the model's
    performance on a specific 3D mesh (e.g., accuracy, robustness, and
    vulnerability to groups of transformations).
    Finally, the aggregate analysis pane extracts insights about the model's
    performance averaged over all the objects and environments collected and
    thus allows the user to notice consistent trends and vulnerabilities in
    their model. 
\end{enumerate}

Each of the aforementioned components (the controls, policy, renderer, inference
module, and logger) are fully customizable and can be extended or replaced 
by the user
without altering the core code of \sandbox{}. For example, while \sandbox{}
supports more than 10 types of controls out-of-the-box, users can add
custom ones (e.g., geometric transformations) by implementing an 
abstract function
that maps a 3D state and a set of parameters  to a 
new state.
Similarly, \sandbox{} supports debugging classification and object detection
models by default, and by implementing a custom evaluator module, users 
can
extend support to a wide variety of other vision tasks and models. 
We refer the reader to Appendix \ref{app:performance} for more information 
on \sandbox{} design principles, implementation, and scalability.

    \section{Debugging and Analyzing Models with \sandbox{}}
    In this section, we illustrate through case studies how to analyze and debug
    vision models with \sandbox{}.
    In each case, we follow the workflow outlined in Section 
    \ref{sec:workflow}---importing
    the relevant objects, selecting the desired transformations (or 
    constructing custom ones), selecting a search policy, and finally
    analyzing the results.

    In all our experiments, we analyze a ResNet-18 \citep{he2015residual}
    trained on the ImageNet \citep{russakovsky2015imagenet}
    classification task (its validation set accuracy is 69.8\%). 
    Note that \sandbox{} is classifier-agnostic (i.e., ResNet-18
    can be replaced with any PyTorch classification module), and even
    supports object detection tasks. For our analysis, we collect 3D 
    models for 16 ImageNet classes (see Appendix
    \ref{app:exp_setup} for more details on each experiment). We ensure that 
    in ``clean'' settings, i.e., when rendered in
    simple poses on a plain white background, the 3D models are correctly
    classified at a reasonable rate (cf. 
    Table~\ref{tab:orig_accs}) by our pre-trained ResNet.

    \begin{table}[h]
        \centering
        {\small
        \begin{tabular}{lcccccccc}
            \toprule
            &  banana &  baseball &  bowl &  drill &  golf ball &  hammer &  lemon &   mug \\
            \midrule
            Simulated accuracy (\%)     &    96.8 &     100.0 &  17.5 &   63.3 &       95.0 &    65.6 &  100.0 &  13.4 \\
            ImageNet accuracy (\%) &    82.0 &      66.0 &  84.0 &   40.0 &       82.0 &    54.0 &   76.0 &  42.0 \\
            \bottomrule
            \toprule
            &  orange &  pitcher base &  power drill &  sandle &  shoe &  spatula &  teapot &  tennis ball \\
            \midrule
            Simulated accuracy (\%)      &    98.5 &           7.9 &         87.5 &    88.0 &  59.2 &     76.1 &    47.8 &        100.0 \\
            ImageNet accuracy (\%) &    72.0 &          52.0 &         40.0 &    66.0 &  82.0 &     18.0 &    80.0 &         68.0 \\
            \bottomrule
        \end{tabular}}
        \caption{Accuracy of a pre-trained ResNet-18, for each of the 16
        ImageNet classes considered, on the corresponding 3D
        model we collected, rendered at an unchallenging pose on a white
        background (``Simulated'' row); and the subset of the ImageNet 
        validation
        set corresponding to the class (``ImageNet'' row).}
        \label{tab:orig_accs}
    \end{table}

    \label{sec:evaluation}
    \subsection{Sensitivity to image backgrounds}
\label{sec:backgrounds_sensitivity}
\begin{figure}[t]
    \centering
    \begin{minipage}{0.32\textwidth}
        \centering
        \includegraphics[trim={0 0 0 0},clip,width=\textwidth]{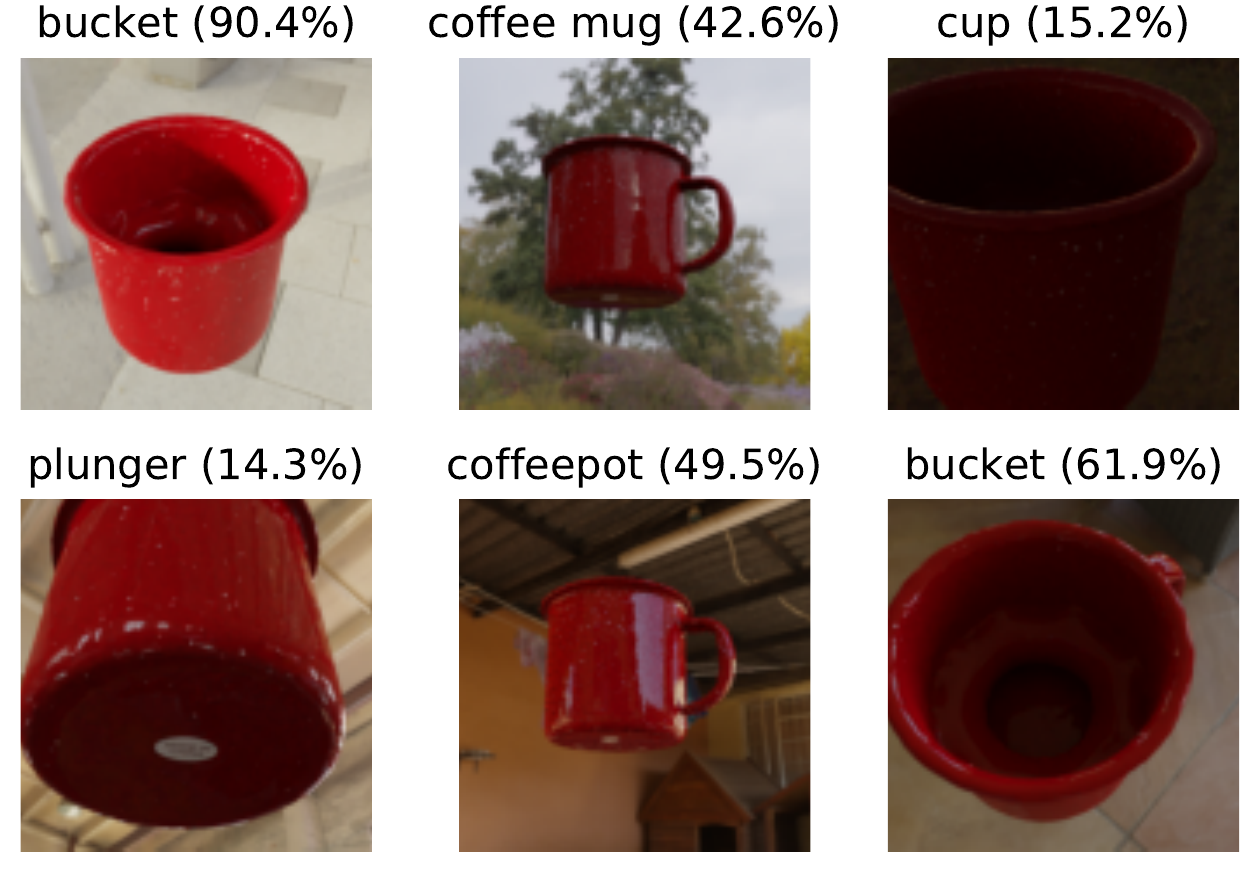}
        \caption{\label{fig:mug_background_samples} Examples of rendered scenes of the coffee
        mug 3D model in different environments,
        labeled with a pre-trained model's top prediction.}
        \label{fig:samples_bg}
    \end{minipage}
    \hspace{1em}
    \begin{minipage}{0.58\textwidth}
        \centering
        \includegraphics[width=1\textwidth]{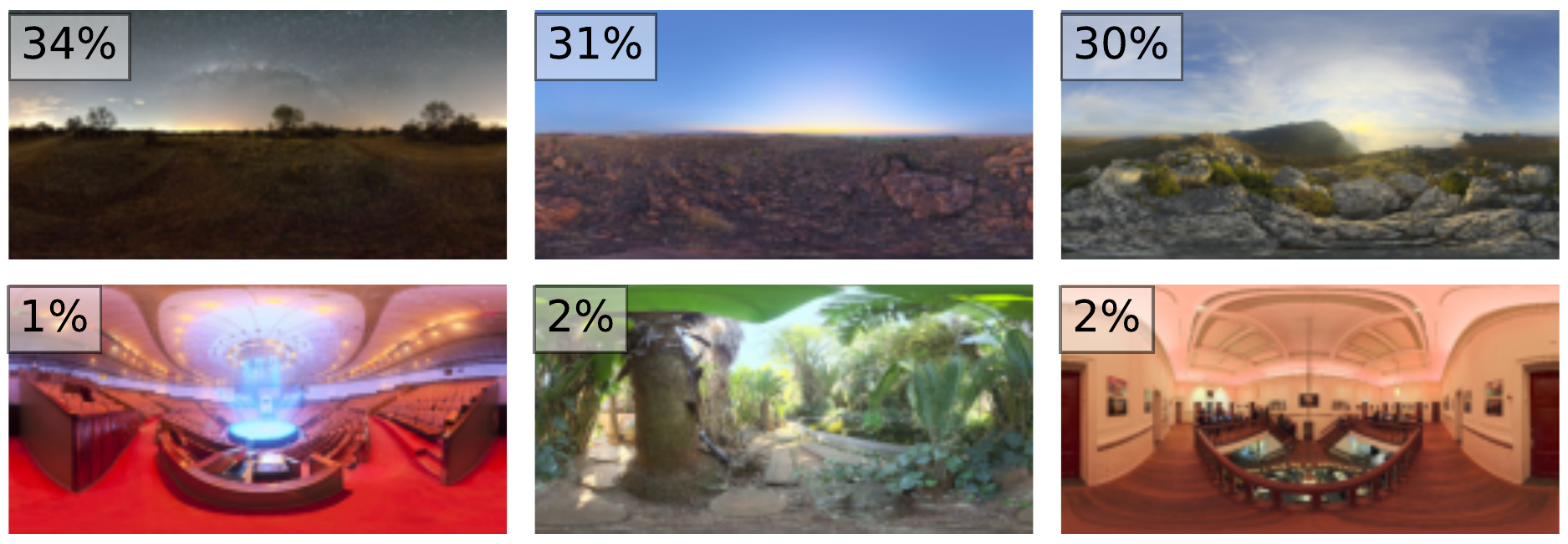}
        \caption{\textbf{(Top)} Best and
        \textbf{(Bottom)} worst background environments for classification of the coffee mug, and
        their respective accuracies (averaged over camera positions and zoom
        factors).} 
        \label{fig:mug_background_best_worst}
    \end{minipage}
\end{figure}

We begin our exploration by using \sandbox{} to confirm ImageNet 
classifiers' reliance on 
background signal, as pinpointed by several recent in-depth
studies~\citep{zhang2007local,zhu2017object,xiao2020noise}. 
Out-of-the-box, \sandbox{} can render 3D models onto HDRI files using
image-based lighting; we downloaded 408 such background environments from 
\url{hdrihaven.com}. 
We then used the pre-packaged ``camera'' and ``orientation''
controls to render (and evaluate our classifier on) scenes of the pre-collected
3D models at random poses, orientations, and scales on each background.
Figure~\ref{fig:samples_bg} shows some (randomly sampled) example scenes 
generated by \sandbox{} for
the ``coffee mug'' model.

\begin{figure}[t]
    \centering
    \includegraphics[trim={0 0 0 0},clip,width=.85\textwidth]{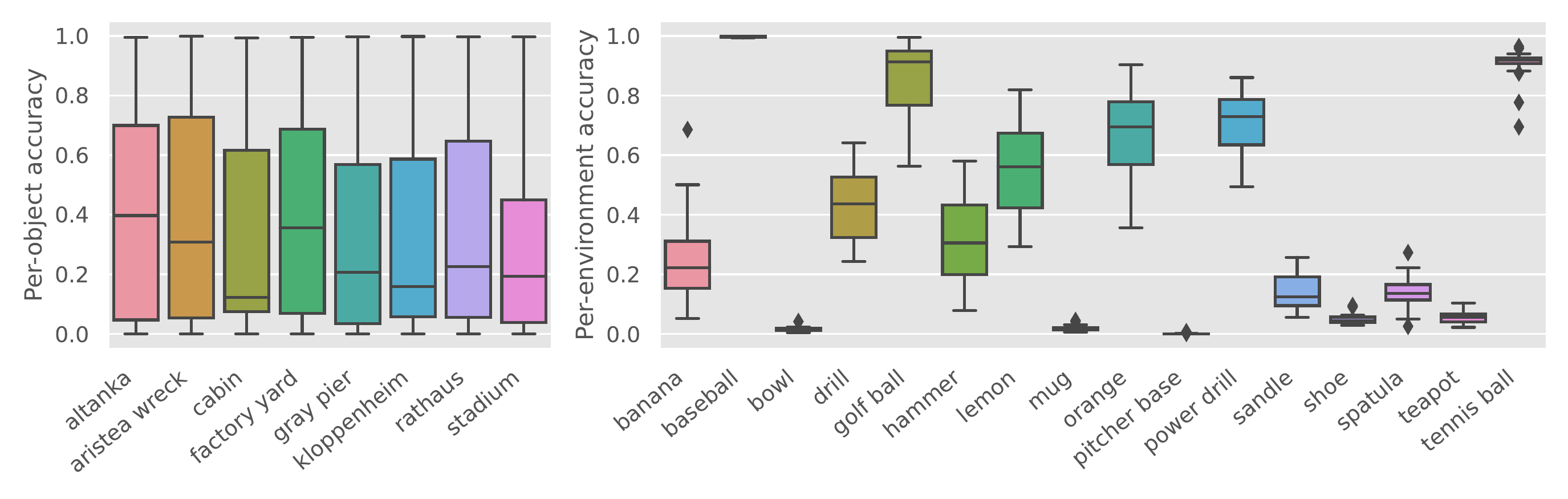}
    \caption{Visualization of accuracy on controls from 
    Section~\ref{sec:backgrounds_sensitivity}.
    \textbf{(Left)} We compute the accuracy of the model conditioned on each object-environment
    pair. For each environment on the x-axis, we plot the variation in accuracy (over the set of possible
    objects) using a boxplot. We visualize the per-object accuracy spread by including the median line,
    the first and third quartiles box edges (the interval between which is
    called the inter-quartile range, IQR), the range, and the outliers
    (points that are outside the IQR by $\nicefrac{3}{2}|\text{IQR}|$).
    \textbf{(Right)} Using the same format, we track how
    the classified object (on the x-axis) impacts variation in accuracy
    (over different environments) on the y-axis.}
    \label{fig:imgnet_bg_dep_per_bg}
\end{figure}

\paragraph{Analyzing a subset of backgrounds.}
In Figure~\ref{fig:imgnet_bg_dep_per_bg}, we visualize the performance of a 
ResNet-18 classifier on the 3D models from 16 different ImageNet 
classes---in random positions, orientations, and scales---rendered onto 
20\footnote{For computational reasons, we subsampled 20 environments 
which we used to analyze all of the pre-collected 3D models.} 
of the collected HDRI backgrounds.
One can observe that background dependence indeed varies widely
across different objects---for example, the ``orange'' and ``lemon'' 3D
models depend much more on background than the ``tennis ball.'' We also find
that certain backgrounds yield systemically higher or lower accuracy; for
example, average accuracy on ``gray pier'' is five times lower than that of
``factory yard.''


\paragraph{Analyzing all backgrounds with the ``coffee mug'' model.} 
The previous study broadly characterizes classifier sensitivity  
classifiers to different models and environments. 
Now, to gain a deeper 
understanding 
of this sensitivity, we focus our analysis only a single 3D model (a ``coffee 
mug'') rendered in all 408 environments.
We find
that the highest-accuracy backgrounds had tags such as \textit{skies},
\textit{field}, and \textit{mountain}, while the lowest-accuracy backgrounds
had tags \textit{indoor}, \textit{city}, and \textit{building}.  

\begin{figure} \centering
    \begin{minipage}{0.48\textwidth}
        \vspace*{-0.3cm}
        \includegraphics[width=\textwidth]{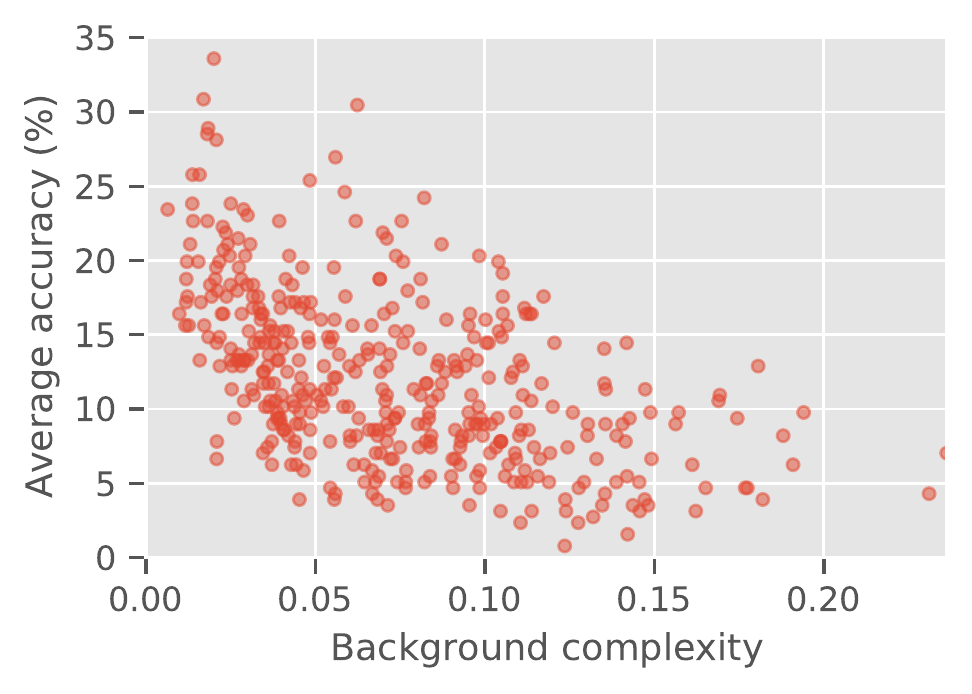}
        \vspace*{-0.5cm}
        \caption{Relation between the complexity of a background and its average accuracy.
        Here complexity is defined as the average pixel value of the image after 
        applying an edge detection
        filter.}
        \label{fig:mug_background_experiment}
    \end{minipage}
    \hspace{1em}
    \begin{minipage}{.38\textwidth}
        \includegraphics[width=\textwidth]{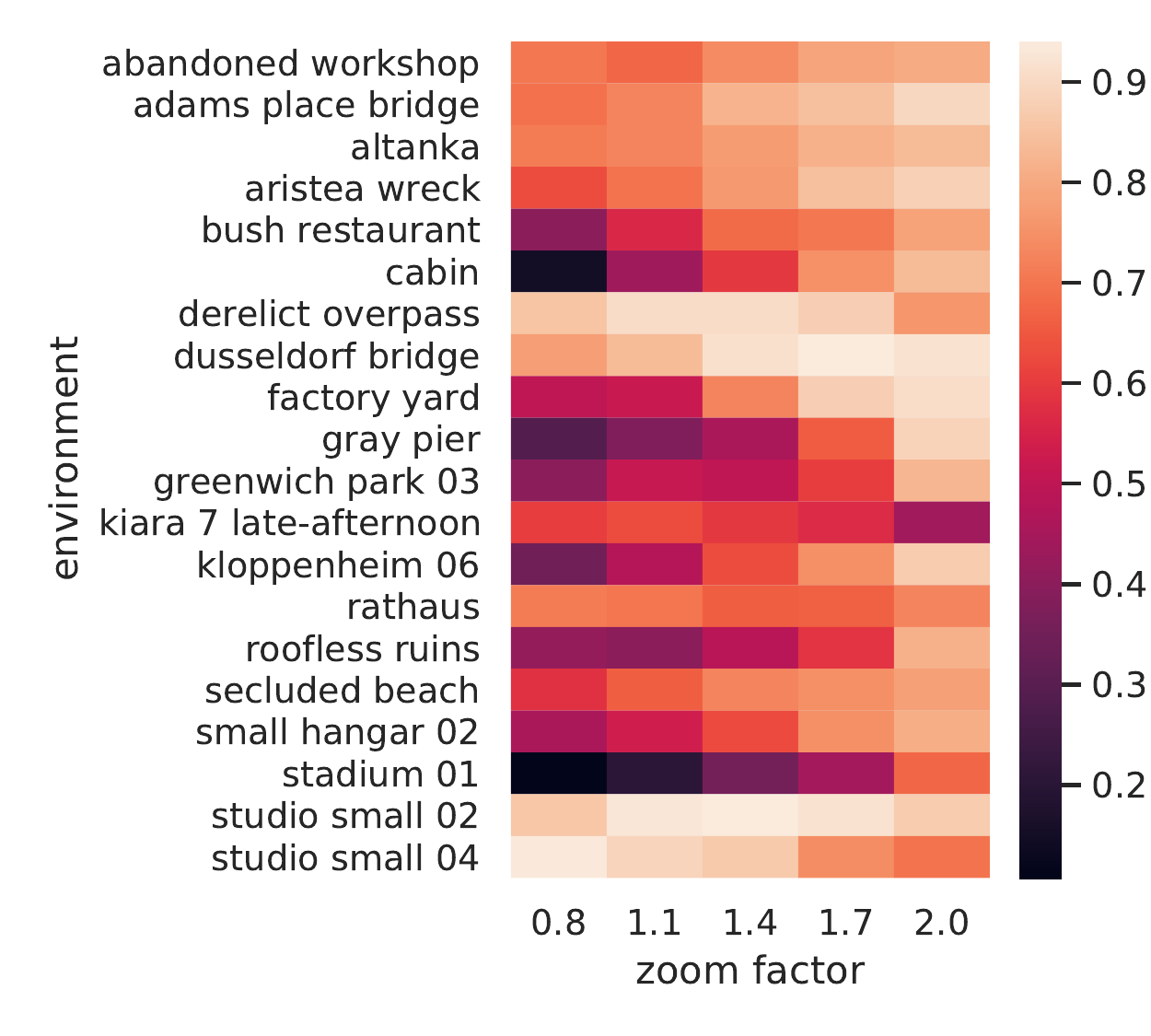}
        \caption{\sandbox{}'s focus on composability enables us to study
        robustness along multiple axes simultaneously. Here we study average 
        model accuracy
        (computed over pose randomization) as a function of {\em both} zoom
        level and background.}  
        \label{fig:orange_xp}
    \end{minipage}
\end{figure}

At first, this observation seems to be at odds with the idea that the classifier 
relies heavily on context clues to make decisions.
After all, the backgrounds where the classifier seems to perform well (poorly) 
are places that we would expect a coffee mug to be rarely (frequently) 
present in the real world.
Visualizing the best and worst backgrounds in terms of accuracy (Figure
\ref{fig:mug_background_best_worst}) suggests a possible explanation for this: 
the best backgrounds tend to be clean and distraction-free. Conversely,
complicated backgrounds (e.g., some indoor scenes) often contain context clues
that make the mug difficult for models to detect.
Comparing a ``background complexity'' metric (based on the number of edges in
the image) to accuracy (Figure \ref{fig:mug_background_experiment}) supports this explanation: mugs overlaid on more complex backgrounds
are more frequently misclassified by the model. 
In fact, some specific backgrounds
even result in the model ``hallucinating'' objects; for example, the second-most
frequent predictions for the \textit{pond} and \textit{sidewalk} backgrounds were  
\textit{birdhouse} and \textit{traffic light} respectively, despite the fact
that neither object is present in the environment.

\paragraph{Zoom/background interactions case study: the advantage of 
composable controls.}
Finally, we leverage \sandbox{}'s composability to study interactions between controls.
In Figure~\ref{fig:orange_xp} we plot the mean
classification accuracy of our ``orange'' model while varying background and 
scale factor.
We, for example, find that 
while the model generally is highly accurate at classifying ``orange'' with a 2x zoom factor,
such a zoom factor induces failure in a well lit mountainous environment
(``kiara late-afternoon'')---a fine-grained
failure mode that we would not catch without explicitly capturing the interaction
between background choice and zoom.

\subsection{Texture-shape bias}
\label{sec:texture-exp}
\begin{figure*}[!htbp]
    \centering
        \begin{minipage}{0.45\textwidth}
            \centering
            \begin{subfigure}{\textwidth}
                \includegraphics[width=\textwidth]{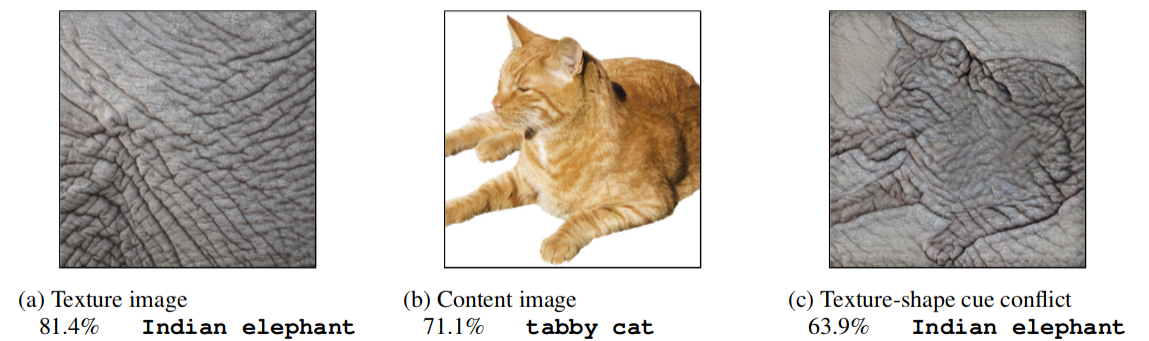}
                \vspace*{0.1em}
            \end{subfigure}
            \begin{subfigure}{\textwidth}
                \centering
                \includegraphics[width=.25\textwidth]{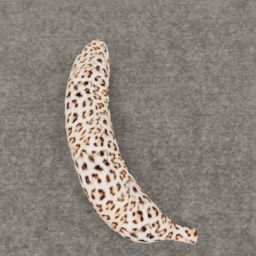}
                \includegraphics[width=.25\textwidth]{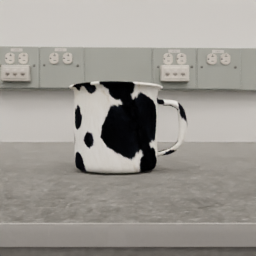}
                \includegraphics[width=.25\textwidth]{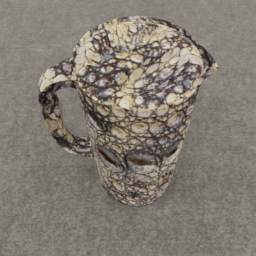}
            \end{subfigure}
            \caption{Cue-conflict images generated by
            \citet{geirhos2018imagenettrained} (\emph{top}) and \sandbox{} 
            (\emph{bottom}).}
        \label{fig:cue_conflict_ex}
        \label{fig:our_cue_conflict}
        \end{minipage}
        \hspace{1em}
        \begin{minipage}{0.45\textwidth}
            \includegraphics[width=\textwidth]{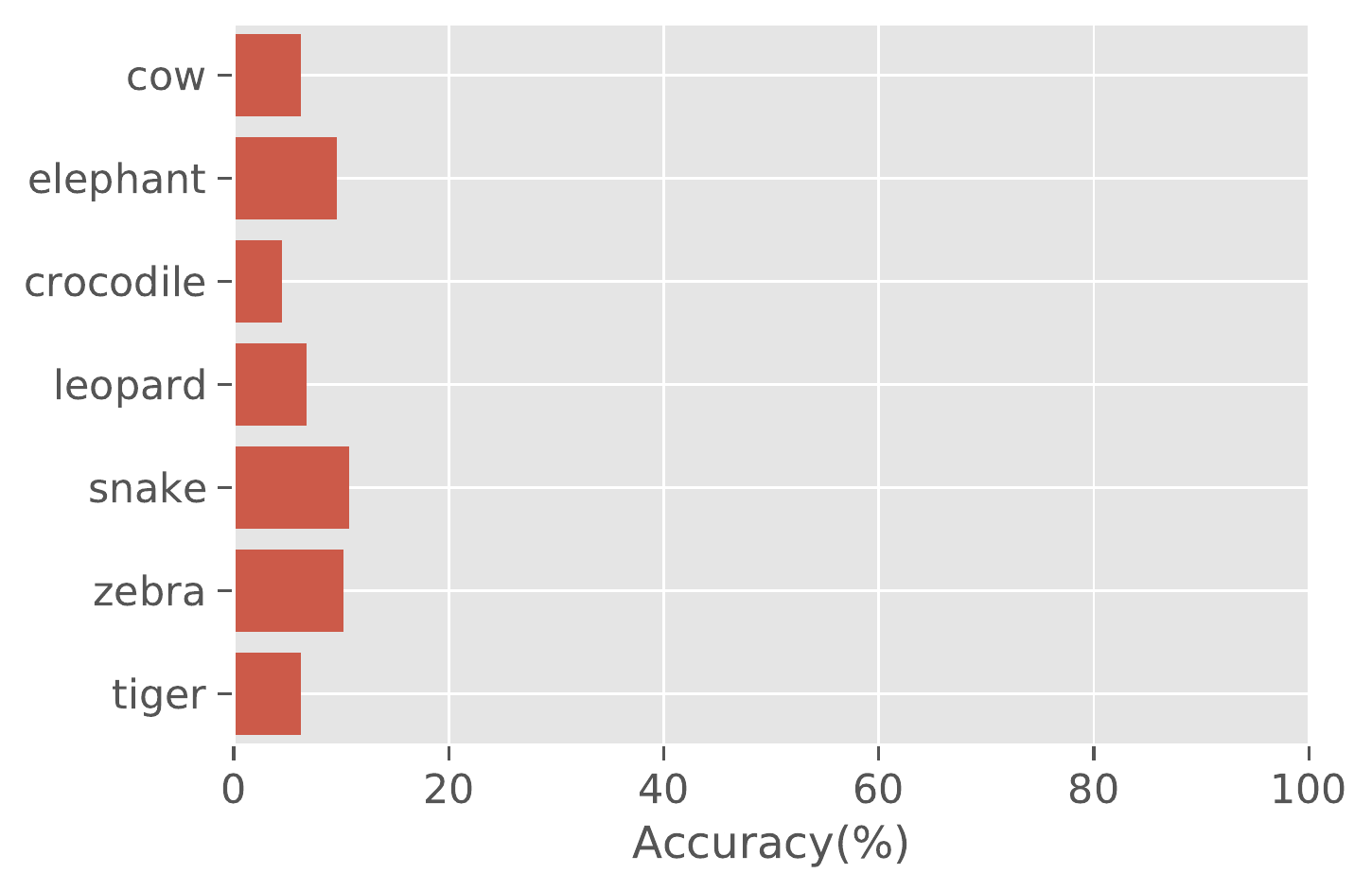}
            \caption{Model accuracy on previously correctly-classified images 
            after their texture is altered via \sandbox{}, as a function of 
            texture-type.}
            \label{fig:texture_swap_experiment_accuracies} 
        \end{minipage}
    \end{figure*}

We now demonstrate how \sandbox{} can be straightforwardly extended to 
discover more complex failure modes in computer vision models.
Specifically, we will show how to rediscover the ``texture bias'' exhibited
by ImageNet-trained neural networks \citep{geirhos2018imagenettrained} in a 
systematic and (near-)photorealistic way.
\citet{geirhos2018imagenettrained} 
fuse pairs of images---combining texture
information from one with shape and edge information from the other---to create
so-called ``cue-conflict'' images.  They then demonstrate that on these images (cf. 
Figure \ref{fig:cue_conflict_ex}), ImageNet-trained convolutional neural networks
(CNNs) typically predict the class corresponding to the texture component, 
while humans typically predict based on shape features.

Cue-conflict images identify a concrete difference between human and CNN
decision mechanisms.  However, the fused images are unrealistic and can be
cumbersome to generate (e.g., even the simplest approach uses style transfer
\citep{gatys2016image}).  \sandbox{} gives us an opportunity to rediscover
the influence of texture in a more streamlined fashion.

Specifically, we implement a control (now pre-packaged with \sandbox{}) that
replaces an object's texture with a random (or user-specified) one.  
We use this control to create cue-conflict objects out of eight 3D
models\footnote{Object models: mug, helmet, hammer, strawberry, teapot, 
pitcher, bowl,
lemon, banana and spatula} and seven animal-skin texture 
images\footnote{Texture types: cow,
crocodile, elephant, leopard, snake, tiger and zebra} (i.e., 56 objects in total).  
We test our pre-trained ResNet-18 on images of these objects rendered in a 
variety of poses and camera locations. Figure
\ref{fig:our_cue_conflict} displays sample cue-conflict images generated 
using \sandbox{}.

\begin{figure}[!htbp]
\centering
\includegraphics[width=\textwidth]{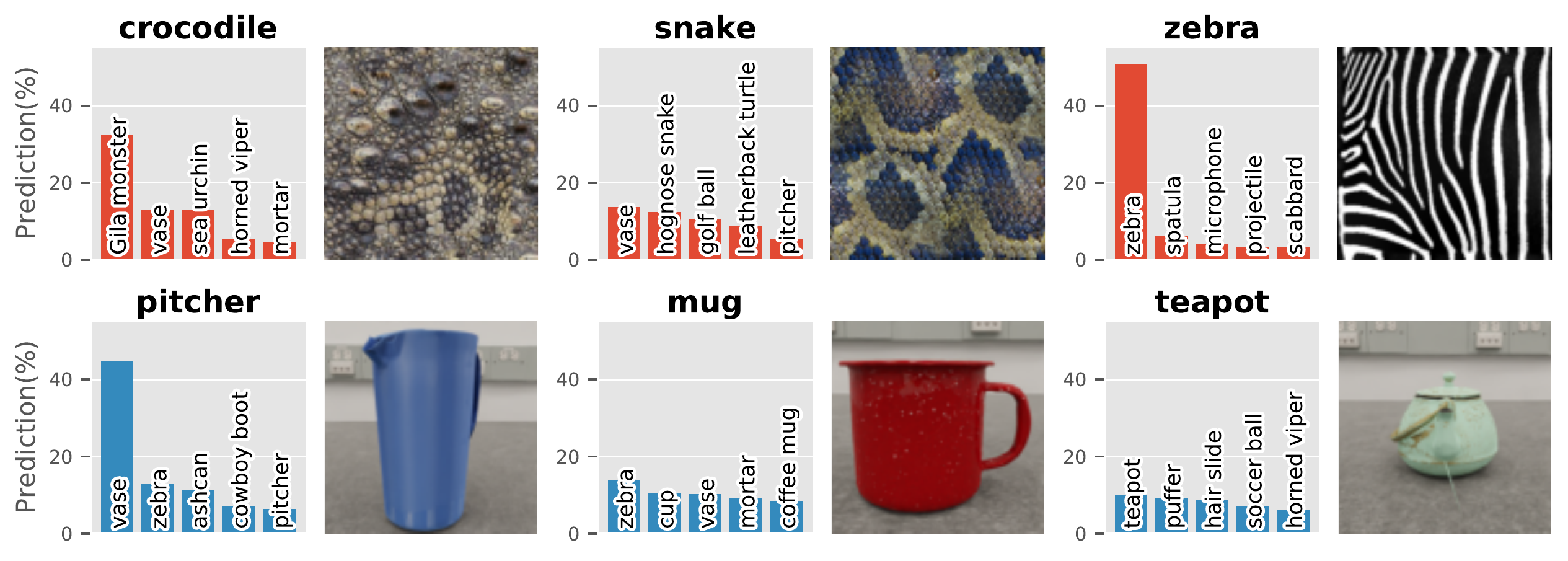}
\caption{\label{fig:texture_swap_histograms_short}
Distribution of classifier predictions after the texture of the 3D object model 
is altered.
In the top row, we visualize the most frequently predicted classes for each
texture (averaged over all objects). In the bottom row, we visualize the most
frequently predicted classes for each object (averaged over all textures). We
find that the model tends to predict based on the texture more often than based
on the object.}
\end{figure}

 Our study confirms the findings of
\citet{geirhos2018imagenettrained} and indicates that texture bias indeed
extends to (near-)realistic settings.  For images that were originally correctly classified
(i.e., when rendered with the original texture), changing the texture reduced
accuracy by 90-95\% uniformly across textures (Figure
\ref{fig:texture_swap_experiment_accuracies}).  Furthermore, we observe that the
model predictions usually align better with the texture of the objects
rather than their geometry  
(Figure~\ref{fig:texture_swap_histograms_short}).  One notable
exception is the pitcher object, for which the most common prediction
(aggregated over all textures) was {\em vase}. A possible explanation for this
(based on inspection of the training data) is that due to high variability of
vase textures in the train set, the classifier was forced to rely more on shape.

\subsection{Orientation and scale dependence}
\label{sec:orientation_scale_dependence}
\begin{figure}
    \includegraphics[width=\textwidth]{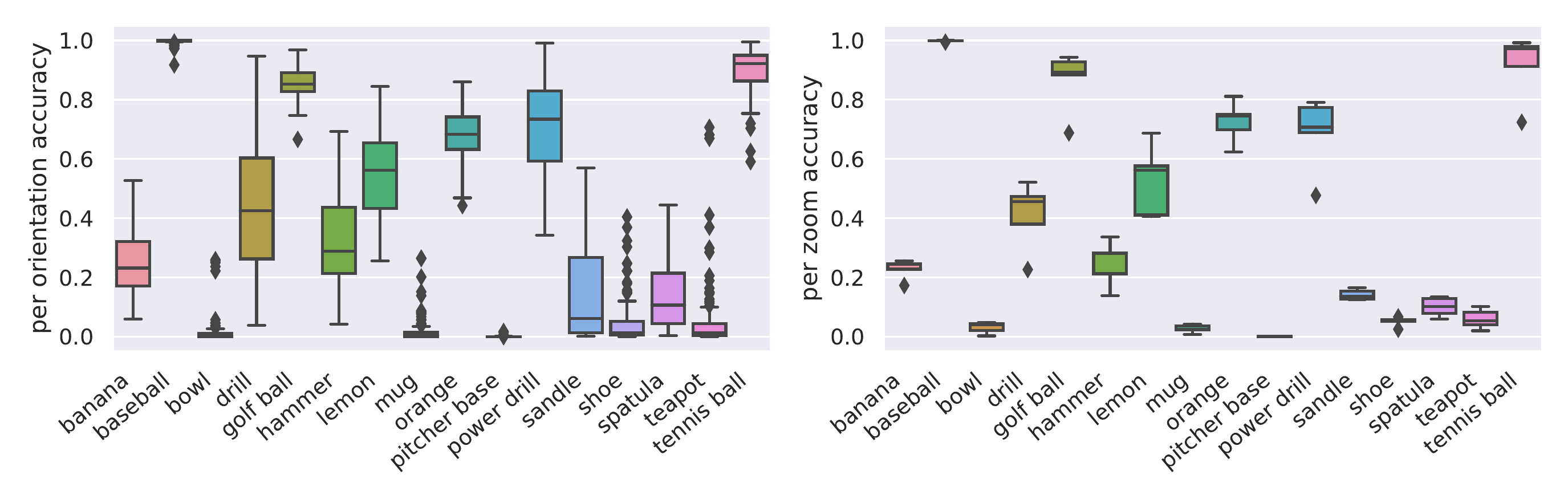}
    \caption{\textbf{(Left)} We compute the accuracy of the model for
        each object-orientation
        pair. For each object on the x-axis, we plot the variation in accuracy
        (over the set of possible orientations) using a boxplot. We visualize
        the per-orientation accuracy
        spread by including the median line,
        the first and third quartiles box edges, the range, and the outliers.
        \textbf{(Right)} Using the same format as the left hand plot, we plot how
        the classified object (on the x-axis) impacts variation in accuracy
        (over different zoom values) on the y-axis.}
    \label{fig:imgnet_orientation_dep}
\end{figure}

Image classification models are brittle to object orientation in both
real and simulated settings~\citep{kanbak2018geometric,engstrom2019rotation,
barbu2019objectnet, alcorn2019strike}. As was the case for both background and
texture sensitivity, reproducing and extending such observations is
straightforward with \sandbox{}. Once again, we use the built-in controls to
render objects at varying poses, orientations, scales, and environments before
stratifying on properties of interest.
Indeed, we find that  classification accuracy
is highly dependent on object orientation
(Figure~\ref{fig:imgnet_orientation_dep} left) and scale (Figure
\ref{fig:imgnet_orientation_dep} right).
However, this dependence is not
uniform across objects. As one would expect, the classifier's accuracy is
 less sensitive to orientation on more symmetric objects
(like ``tennis ball'' or ``baseball''), but can vary widely on more uneven
objects (like ``drill'').

\begin{figure}
    \includegraphics[width=0.15\textwidth]{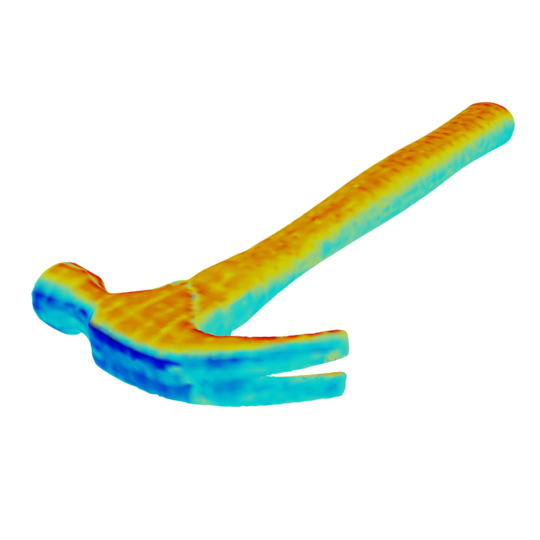}
    \includegraphics[width=0.15\textwidth]{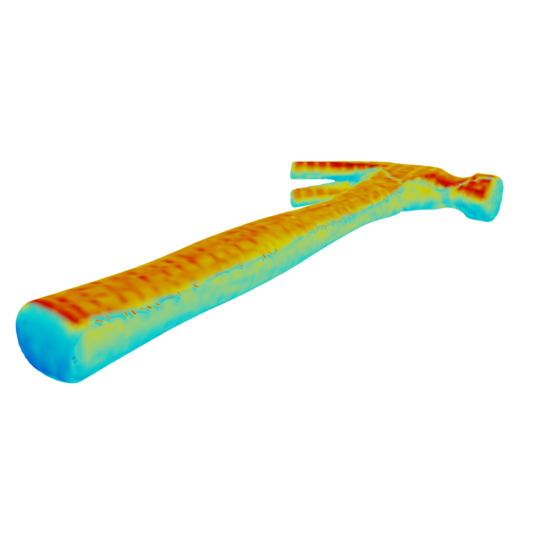}
    \hspace{1em}
    \includegraphics[width=0.15\textwidth]{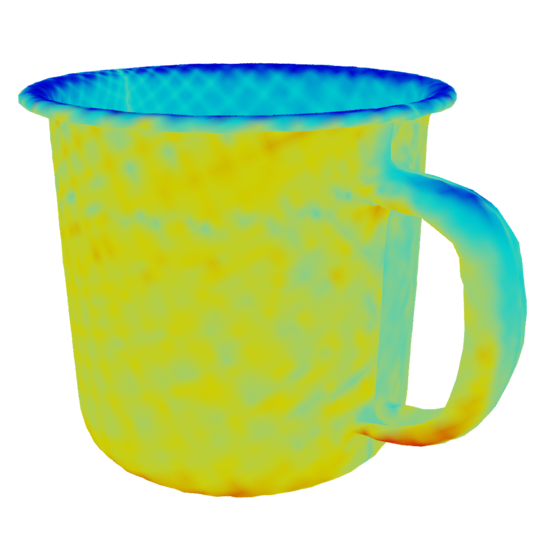}
    \includegraphics[width=0.15\textwidth]{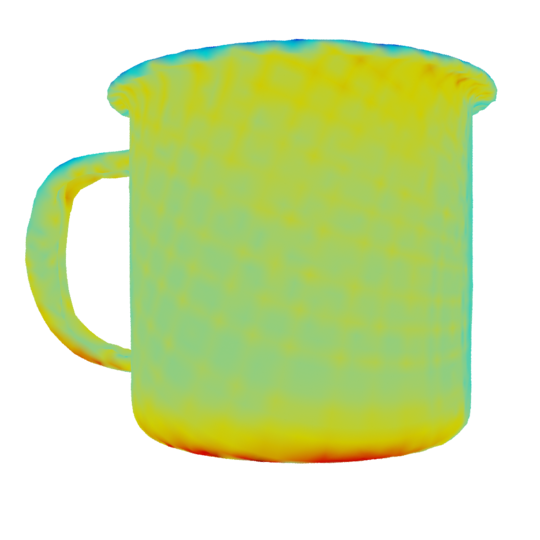} 
    \hspace{1em}
    \includegraphics[width=0.15\textwidth]{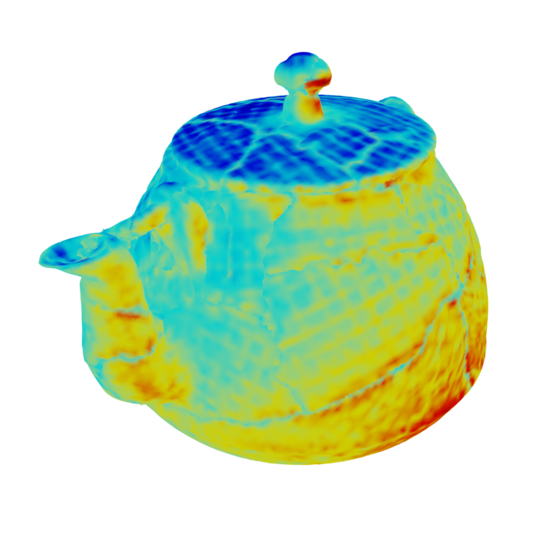}
    \includegraphics[width=0.15\textwidth]{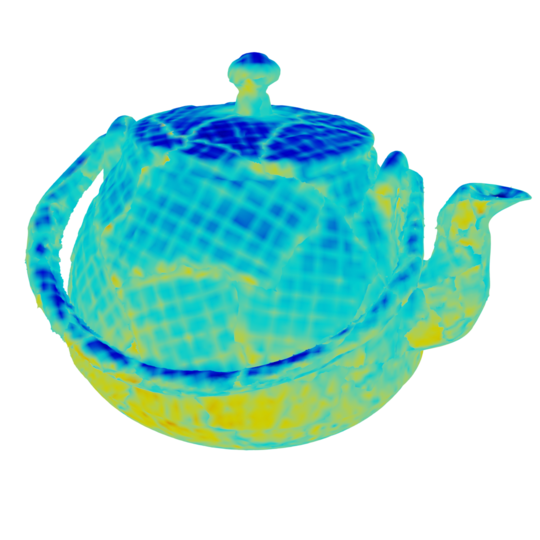}
    \caption{Model sensitivity to pose. The heatmaps denote the 
    	accuracy of the model in predicting the correct label, conditioned on a 
    	specific part of the object being visible in the image. Here, red and blue 
    	denotes high and low accuracy respectively.}
    \label{fig:heatmap_visualizations}
\end{figure}

For a more fine-grained look at the importance of object orientation, we can
measure the classifier accuracy conditioned on a given part of each 3D model
being visible. This analysis is once again straightforward in \sandbox{}, since
each rendering is (optionally) accompanied by a UV map which maps pixels in the
scene back to locations on on the object surface. Combining these UV maps with
accuracy data allows one to construct the ``accuracy heatmaps'' shown in Figure 
\ref{fig:heatmap_visualizations}, wherein each part of an object's surface corresponds to
classifier accuracy on renderings in which the part is visible. The results confirm
that atypical viewpoints adversely impact model performance, and also allow
users to draw up a variety of testable hypotheses regarding performance on
specific 3D models (e.g., for the coffee mug, the bottom rim is highlighted in
red---is it the case that mugs are more accurately classified when viewed from
the bottom)? These hypotheses can then be investigated further through natural
data collection, or---as we discuss in the upcoming section---through additional
experimentation with \sandbox{}.

\subsection{Case study: using \sandbox{} to dive deeper}
\label{subsec:mug_fluid}
Our heatmap analysis in the previous section (cf. Figure
\ref{fig:heatmap_visualizations}) showed that classification accuracy
for the mug decreases when its interior is visible. 
What could be causing this effect? 
One hypothesis is that in the ImageNet training set, objects are captured in
context, and thus ImageNet-trained classifiers rely on this context to make
decisions. Inspecting the ImageNet dataset, we notice that coffee mugs in 
context usually contain coffee in them.
Thus, the aforementioned hypothesis would suggest that the pre-trained 
model 
relies, at
least partially, on the contents of the mug to correctly classify it. 
\textit{Can we leverage \sandbox{} to confirm or refute this hypothesis?}

To test this, we implement a custom control that can render a liquid inside 
the ``coffee mug'' model.
Specifically, this control takes water:milk:coffee ratios as parameters, then
uses a parametric Blender shader (cf. Appendix \ref{app:omitted_figures})
to render a corresponding mixture of the liquids into the mug.
We used the pre-packaged grid search policy, (programmatically) restricting 
the 
search space to viewpoints from which the interior of the mug was visible.

\begin{figure}
\centering
\begin{subfigure}{.35\textwidth}
\centering
\vspace*{-0.2cm}
\includegraphics[width=\textwidth]{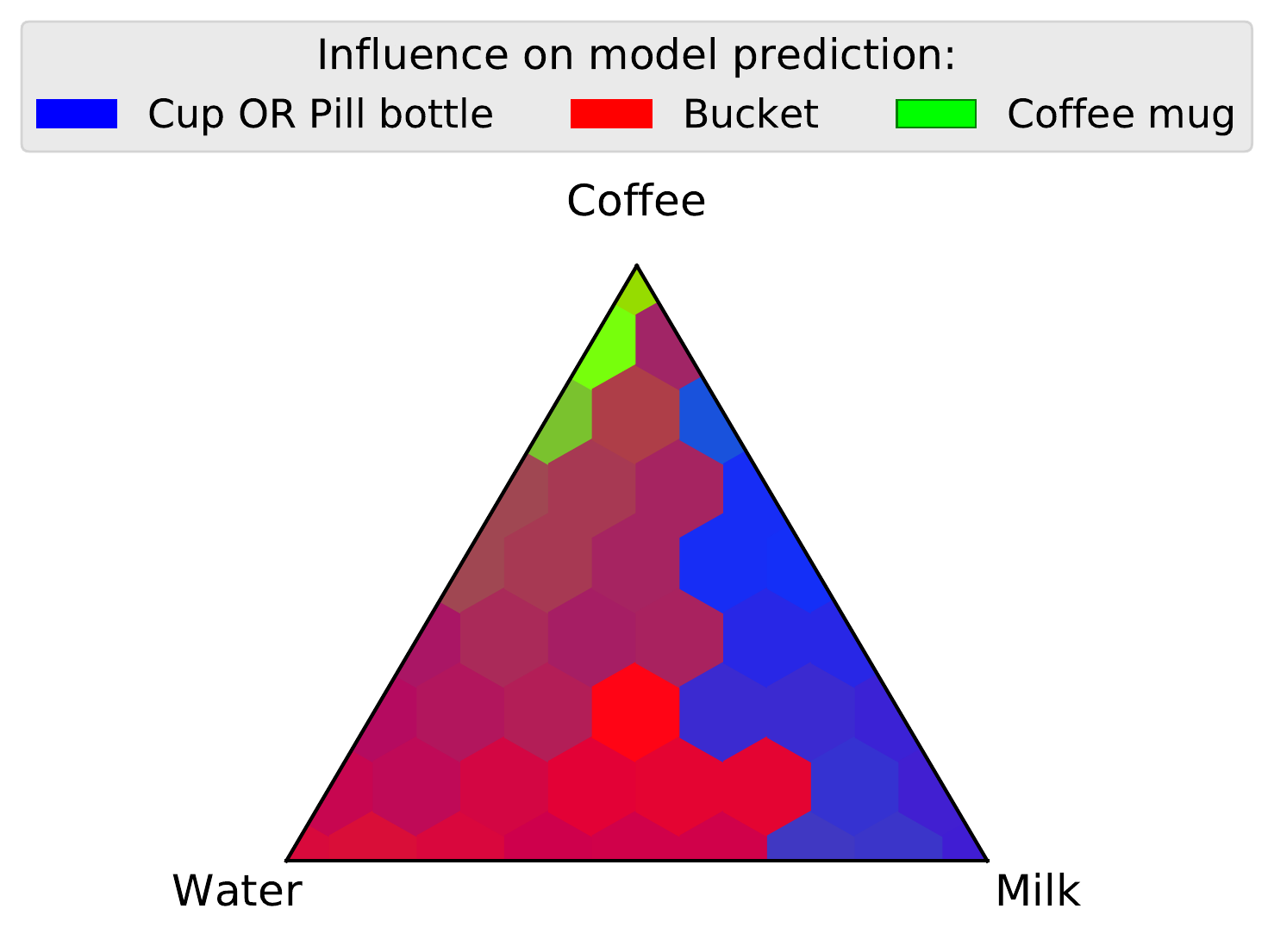}
\caption{}
\label{fig:mug_liquid_experiment_simplex}
\end{subfigure}
\begin{subfigure}{.35\textwidth}
\centering
\includegraphics[width=\textwidth]{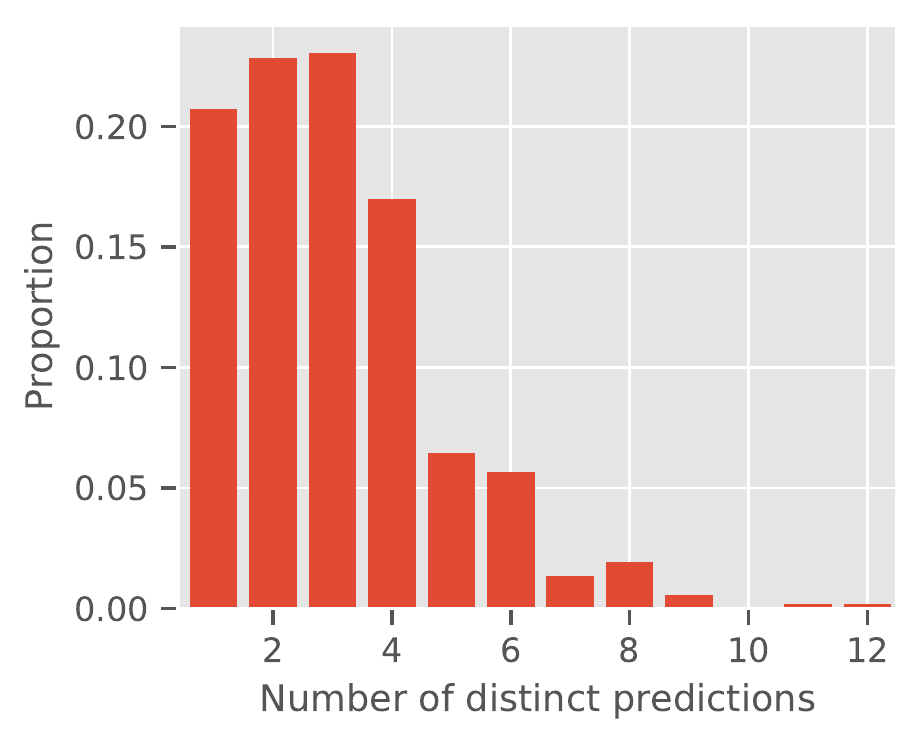}
\caption{}
\label{fig:mug_liquid_experiment_consistency}
\end{subfigure}
\begin{subfigure}{.23\textwidth}
\centering
\vspace*{-0.6cm}
\includegraphics[width=\textwidth]{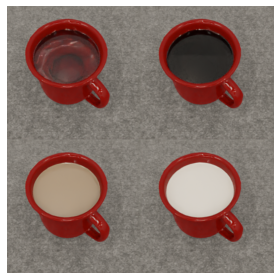}
\caption{}
\label{fig:mug_liquid_experiment_samples}
\end{subfigure}
\caption{Testing classifier sensitivity to context: Figure (a) shows the
correlation of the liquid mixture in the mug on the prediction of the
model, averaged over random viewpoints (see Figure
\ref{fig:mug_liquid_experiment_simplex_raw} for the raw frequencies). Figure (b)
shows that for a fixed viewpoint, model predictions are unstable with respect to
the liquid mixture. Figure (c) shows examples of rendered liquids (water, black
coffee, milk, and milk/coffee mix).}
\label{fig:mug_fluid_results}
\end{figure}

The results of the experiment are shown in Figure \ref{fig:mug_fluid_results}.
It turns out that the model is indeed sensitive to changes in liquid, supporting our
hypothesis: model predictions stayed constant (over all liquids) for only 20.7\%
of the rendered viewpoints (cf. Figure \ref{fig:mug_liquid_experiment_consistency}).
The \sandbox{} experiment provides further support for the hypothesis when we
look at the correlation between the liquid mixture and the predicted class:
Figure \ref{fig:mug_liquid_experiment_simplex} visualizes this correlation in a
normalized heatmap (for the unnormalized version, see Figure
\ref{fig:mug_liquid_experiment_simplex_raw} in the Appendix~\ref{app:omitted_figures}).
We find that the model is most likely to predict ``coffee mug'' when coffee
is added to the interior (unsurprisingly); as the coffee is mixed with water or
milk, the predicted label distribution shifts towards ``bucket'' and ``cup'' or
``pill bottle,'' respectively.
Overall, our experiment suggests that current ResNet-18 classifiers are 
indeed sensitive to object context---in this case, the fluid composition of the 
mug interior. 
More broadly, this illustration highlights how a system designer can quickly
go from hypothesis to empirical verification with minimal effort using \sandbox{}. 
(In fact, going from the initial hypothesis to Figure
\ref{fig:mug_fluid_results} took less than a single day of work for one author.)

\section{Physical realism}
\label{sec:realworld-exp}
The previous sections have demonstrated various ways in which we can use
\sandbox{} to obtain insights into model behavior in simulation.  
Our overarching goal, however, is to understand when models will fail in the
physical world. 
Thus, we would like for the insights extracted by \sandbox{}
to correspond to naturally-arising model behavior, and not just artifacts of
the simulation itself
\footnote{Indeed, a related challenge is the {\em sim2real}
problem in reinforcement learning, where agents trained in
simulation latch on to simulator properties and fail to generalize to the
real world. In both cases, we are concerned about artifacts or spurious
correlations that invalidate conclusions made in simulation.}. 
To this end, we now test the {\em physical realism} of
\sandbox{}: can we understand model performance (and uncover vulnerabilities)
on real photos using only a high-fidelity simulation?

To answer this question, we collected a set of physical objects with
corresponding 3D models, and set up a physical room with its corresponding
3D environment.
We used \sandbox{} to identify strong points and vulnerabilities of a 
pre-trained ImageNet classifier in this environment, mirroring our
methodology from Section~\ref{sec:evaluation}. 
We then recreated each scenario found by \sandbox{} in the physical room,
and took photographs that matched the simulation as closely as possible. 
Finally, we evaluated the physical realism of the system by comparing models'
performance on the photos (i.e., whether they classified each photo correctly)
to what \sandbox{} predicted. 

\paragraph{Setup.} 
We performed the experiment in the studio room shown in Appendix 
Figure~\ref{fig:msr_studio_real} for which we obtained a fairly accurate
3D model (cf. Appendix Figure~\ref{fig:msr_studio_synth}).
We leverage the \texttt{YCB} \cite{calli2015benchmarking} dataset to guide 
our selection of real-world objects, for which 3D models are available. 
We supplement these by sourcing additional objects (from 
\texttt{amazon.com}) and using a 3D scanner to obtain
corresponding meshes. 
\footnote{We manually adjusted the textures of these 3D 
models to increase realism (e.g., by
	tuning reflectance or roughness). In particular, classic photogrammetry is 
	unable to 
model the metallicness and reflectivity of objects. It also tends to embed
reflections as part of the color of the object} 

We next used \sandbox{} to analyze the performance of a pre-trained ImageNet
ResNet-18 on the collected objects in simulation, varying over a set of
realistic object poses, locations, and orientations. 
For each object, we selected 10 rendered situations: five where the model
made the correct prediction, and five where the model predicted incorrectly.
We then tried to recreate each rendering in the physical world. First we roughly
placed the main object in the location and orientation specified in the
rendering, then we used a custom-built iOS application (see Appendix~\ref{app:iphone_app}) to more precisely match the rendering with the physical
setup.

\begin{figure}
    \centering
    \hspace*{0.1cm}
    \includegraphics[width=0.98\textwidth]{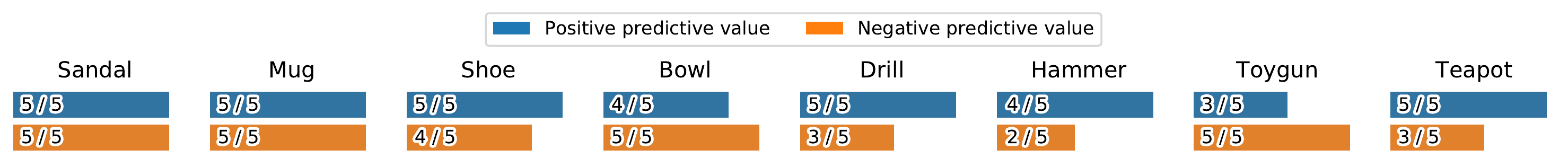}
    \includegraphics[width=\textwidth]{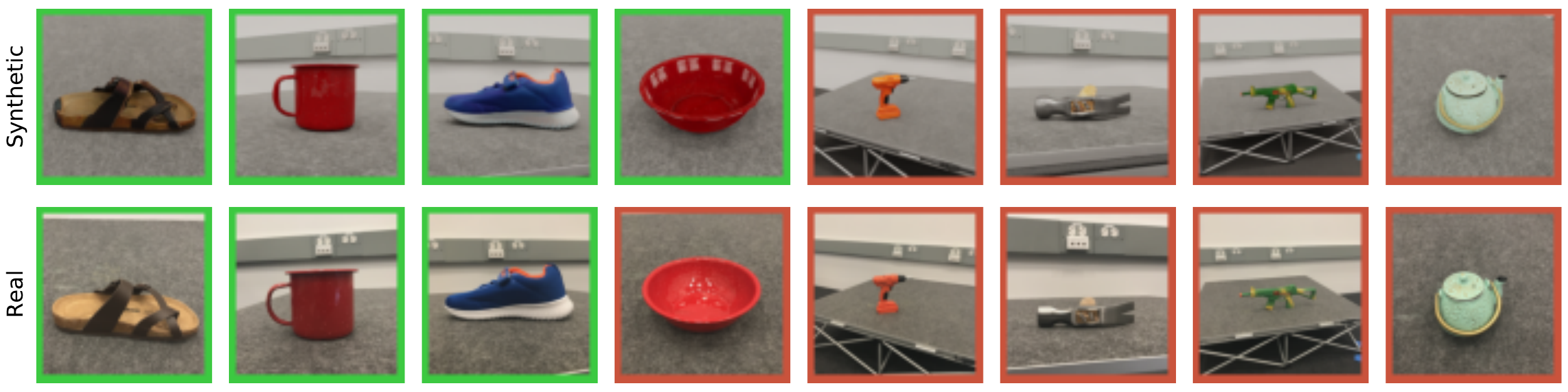}
    \caption{{\bf (Top)} Agreement, in terms of model correctness, between model
    predictions within \sandbox{} and model predictions in the real world. 
    For each object, we selected five rendered scenes
    found by \sandbox{} that were misclassified in simulation, and five that
    were correctly classified; we recreated and deployed the model on each scene
    in the physical world.  
    The \textit{positive (resp., negative) predictive
    value} is rate at which correctly (resp. incorrectly) classified examples in
    simulation were also correctly (resp., incorrectly) classified in the
    physical world. 
    {\bf (Bottom)} Comparison between example simulated scenes generated by \sandbox{} (first
    row) and their recreated physical counterparts (second row). Border color
    indicates whether the model was correct on this specific image.}
    \label{fig:real_life_exp_samples}
\end{figure}

\paragraph{Results.} Figure \ref{fig:real_life_exp_samples} visualizes a few
samples of renderings with their recreated physical counterparts, annotated with
model correctness.
Overall, we found a 85\% agreement rate between the model's
correctness on the real photos and the synthetic renderings---agreement rates
per class are shown in Figure \ref{fig:real_life_exp_samples}. 
Thus, despite imperfections in our physical reconstructions, the vulnerabilities
identified by \sandbox{} turned out to be physically realizable
vulnerabilities (and conversely, the positive examples found by \sandbox{} are
usually also classified correctly in the real world).
We found that objects with simpler/non-metallic materials (e.g., the bowl,
mug, and sandal) tended to be more reliable than metallic objects such as the
hammer and drill. It is thus possible that more precise texture tuning of 3D 
models object could increase agreement further (although a more 
comprehensive study would be needed to verify this).

    \label{sec:extensibility}
    \section{Extensibility}

\begin{figure}
    \begin{subfigure}[b]{0.33\textwidth}
        \centering
        \includegraphics[width=0.47\textwidth]{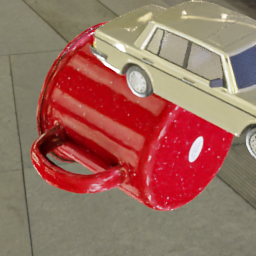}
        \includegraphics[width=0.47\textwidth]{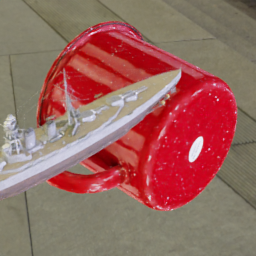}
        \caption{Occlusion control}
        \label{fig:occlusion_control}
    \end{subfigure}
    \hfill
    \begin{subfigure}[b]{0.33\textwidth}
        \centering
        \includegraphics[width=0.47\textwidth]{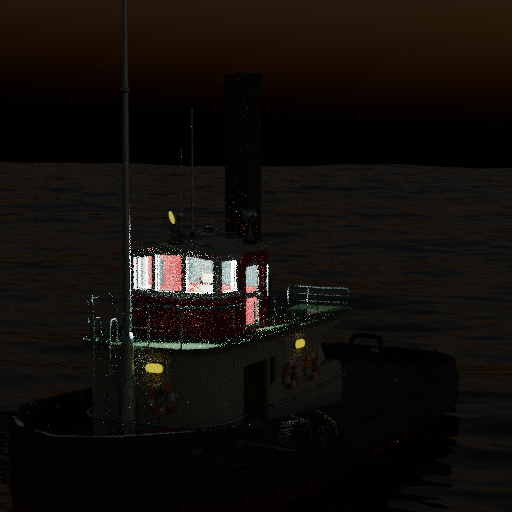}
        \includegraphics[width=0.47\textwidth]{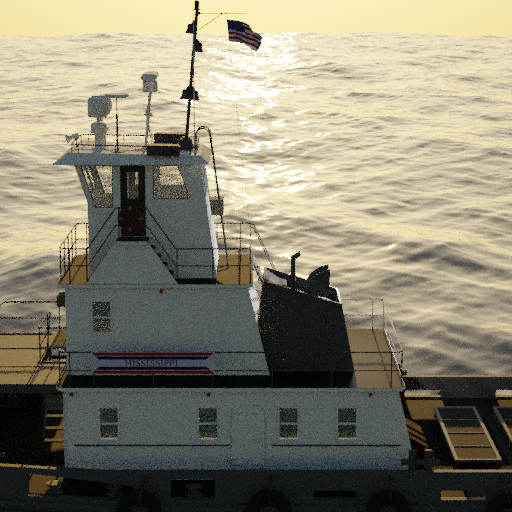}
        \caption{``Time of day'' control}
        \label{fig:time_of_day_control}
    \end{subfigure}
    \begin{subfigure}[b]{0.33\textwidth}
        \centering
        \includegraphics[width=0.94\textwidth]{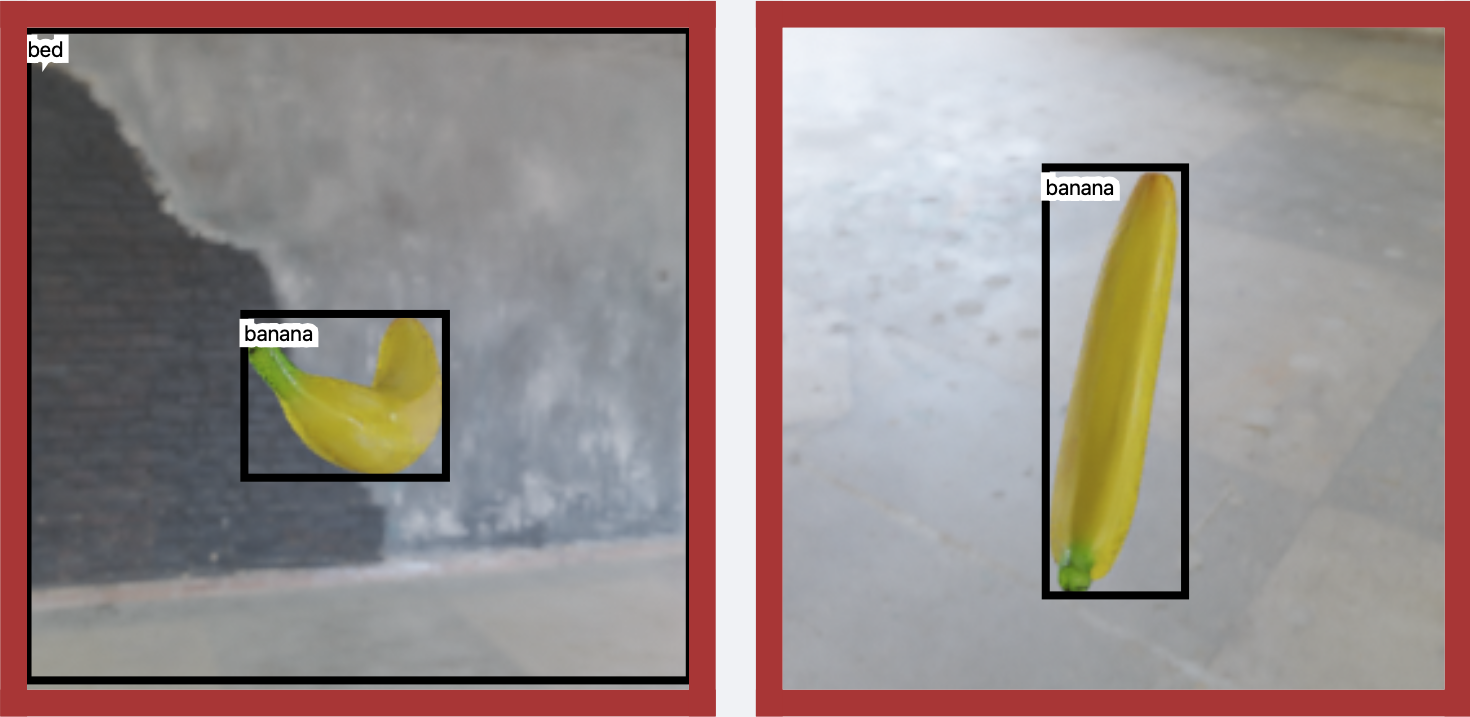}
        \caption{Object detection custom objective}
        \label{fig:detection}
    \end{subfigure}
    \caption{\label{fig:extra_controls} Example of some of the ways in which one
    can extend \sandbox{}: adding custom controls, defining custom objectives,
    and integrating external libraries.}
\end{figure}

\sandbox{} was designed with extensibility in mind. Indeed, the behavior of {\em
every} component of the framework can be substituted with other
(built-in, third-party, or custom-made) implementation.
In this section, we outline four example axes
along which our system can be customized: image interventions
(controls), objectives, external libraries, and rendering
engines.
Our documentation~\cite{threedbdocumentation} provides further details and
step-by-step tutorials.

\paragraph{Custom controls.}
As we have discussed in the previous sections, there is a large body of work
studying the effects of input transformations on model
predictions~\cite{xiao2020noise,liu2018dpatch,rosenfeld2018elephant,geirhos2018imagenettrained,zhu2017object,wang2017afastrcnn}.
The input interventions that these works utilized included, for example, separating foregrounds from
backgrounds \cite{xiao2020noise,zhu2017object}, adding overlays on top of
images \cite{liu2018dpatch,rosenfeld2018elephant,wang2017afastrcnn}, and performing
style transfer \cite{geirhos2018imagenettrained}. These interventions have been implemented with a lot of care. However, they still tend to introduce artifacts and
can lack realism. In Section \ref{sec:evaluation} we already demonstrated that \sandbox{} is
able to circumvent these problems in a streamlined and 
composable manner. Indeed, by operating in three dimensional space, i.e., 
before rendering happens, \sandbox{} enables image transformations that are less
labor-intensive to implement and produce more realistic outputs.
To showcase this, in Section \ref{sec:evaluation} we replicated various image transformation
studies using the {\em controls} built in to \sandbox{} (e.g., Figure
\ref{fig:texture_swap_experiment_accuracies} corresponds to the study of 
\citep{geirhos2018imagenettrained}). However, beyond these built-in capabilities, users
can also add custom controls that implement their desired transformations:
Figure \ref{fig:occlusion_control}, for example, depicts the output of a custom
``occlusion control'' that could be used to replicate studies such as
\cite{rosenfeld2018elephant}.

\paragraph{Custom objectives.}
Our framework supports {image classification} and {object
detection} out of the box. (In this paper, we focus primarily on the former---cf.
Figure \ref{fig:detection} for an example of the latter.)
Still, users can extend \sandbox{} to imbue it with an ability to analyze models for a wide
variety of vision tasks.
In particular, 
in addition to the images shown throughout this work, \sandbox{} renders (and provides
an API for accessing) the corresponding segmentation and depth maps. 
This allow users to easily use the framework for tasks such as {depth estimation},
{instance segmentation}, and
{image segmentation} (the last one of these is in fact subject of our tutorial on
the implementation of custom tasks\footnote{\url{https://3db.github.io/3db/usage/custom_evaluator.html}}). 
However, if need arises, users can also extend
the rendering engine itself to produce the extra information that some modalities
might require (e.g., the coordinates of joints for {pose estimation}).

\paragraph{External libraries.}
\sandbox{} also streamlines the incorporation of external libraries for image
transformations. For example, the ImageNet-C~\cite{hendrycks2019benchmarking} 
corruptions can be integrated into a \sandbox{} control pipeline with very little
effort. (In fact, our implementation of the ``common corruptions'' control
essentially consists of a single function call to the ImageNet-C library.)

\paragraph{Rendering engine.}
Blender~\cite{blender}, the default rendering backend for \sandbox{}, offers a broad set
of features. Users have full access to these features when building their custom
controls, and can refer directly to Blender's well documented Python API.
To illustrate that fact, we leveraged one
of Blender's procedural sky models (\cite{nishita1993,hosek2013,preetham1999}) to
implement a {control} that simulates illumination at different times
of the day (cf. Figure \ref{fig:time_of_day_control}). 

We selected Blender as the backend for \sandbox{} due to the way it balances
ease of use, fidelity, and performance. 
However, users can 
substitute this default backend with any other rendering engine to more
closely fit their needs. For example, users can, on the one hand, setup a rendering backend
(and corresponding controls) based on Mitsuba \cite{mitsuba}, a
research-oriented engine capable of highly accurate simulation.
On the other hand, they can achieve real-time performance at the expense of
realism by implementing a custom backend using a rasterization engine such as
Pandas3D~\cite{pandas3d}.

    \section{Related Work}
    \label{sec:related}
    
\sandbox{} builds on a growing body of
work that looks beyond accuracy-based benchmarks in order to understand the {\em
robustness} of modern computer vision models and their failure modes. In
particular, our goal is to provide a unified framework for reproducing these
studies and for conducting new such analyses. In this section, we discuss the existing research in robustness, interpretability, and
simulation that provide the context for our work.

\paragraph{Adversarial robustness.} Several recent works propose analyzing model
robustness by crafting adversarial, i.e., {\em worst-case}, inputs. For example,
\citep{szegedy2014intriguing} discovered that a carefully chosen but
imperceptible perturbation suffices to change classifier predictions on
virtually any natural input. Subsequently, the study of such ``adversarial
examples'' has extended far beyond the domain of image classification: e.g.,
recent works have studied worst-case inputs for object detection and image
segmentation 
\citep{eykholt2018physical,xie2017adversarial,fischer2017adversarial};
generative models \citep{kos2017adversarial}; and reinforcement learning
\citep{huang2017adversarial}. More closely related to our work are studies
focused on three-dimensional or physical-world adversarial examples
\citep{eykholt2018physical, brown2018adversarial,
athalye2018synthesizing, xiao2019meshadv, LiuTLNJ2019ICLR}. These studies 
typically use differentiable rendering and perturb object texture,
geometry, or lighting to induce misclassification. 
Alternatively, \citet{li2019adversarial} modify the camera itself via an
adversarial camera lens that consistently cause models to misclassify inputs.

In our work, we have primarily focused on using non-differentiable but
high-fidelity rendering to analyze a more {\em average-case} notion of model
robustness to semantic properties such as object orientation or image
backgrounds. Nevertheless, the extensibility of \sandbox{} means that users can
reproduce such studies (by swapping out the Blender rendering module for a
differentiable renderer, writing a custom control, and designing a custom search
policy) and use our framework to attain a more realistic understanding of the
worst-case robustness of vision models. 

\paragraph{Robustness to synthetic perturbations.} 
Another popular approach to analyzing
model robustness involves applying transformations to natural images and 
measuring the resultant changes in model predictions. For example,
\citet{engstrom2019rotation} measures robustness to image rotations and
translations; \citet{geirhos2018imagenettrained} study robustness to style
transfer (i.e., texture perturbations); and a number of works has
studied robustness to common corruptions
\citep{hendrycks2019benchmarking,kang2019testing}, changes in image backgrounds
\citep{zhu2017object, xiao2020noise}, Gaussian noise
\citep{ford2019adversarial}, and object occlusions
\citep{rosenfeld2018elephant}, among other transformations.

A more closely related approach to ours analyzes the impact of factors such as
object pose and geometry by applying synthetic perturbations in
three-dimensional space \citep{HamdiG2019, ShuLQY2020, HamdiMG2018,
alcorn2019strike}. For example, \citet{HamdiG2019} and \citet{JainCJWLYCJS2020}
use a neural mesh renderer \cite{KatoUH2018meshrenderer} and Redner 
\cite{LiADL2018redner} respectively to render images to analyze the failure 
modes of vision models. \citet{alcorn2019strike} present a system for 
discovering neural networks failure modes as a function of object orientation,
zoom, and (two-dimensional) background and perform a thorough study on the
impact of these factors on model decisions.

\sandbox{} draws inspiration from the studies listed above and
tries to provide a unified framework for detecting {\em arbitrary} model failure
modes. For example, our framework provides explicit mechanisms for users to make
custom controls and custom search strategies, and includes built-in controls
designed to range many possible failure modes encompassing nearly all of the
aforementioned studies (cf. Section \ref{sec:evaluation}). Users can also {\em
compose} different transformations in \sandbox{} to get an even more
fine-grained understanding of model robustness.

\paragraph{Other types of robustness.} An oft-studied but less related branch of
robustness research tests model performance on unaltered images from
distributions that are nearby but non-identical to that of the training
set. Examples of such investigations include studies of newly collected datasets
such as ImageNet-v2
\citep{recht2018imagenet,engstrom2020identifying,taori2020measuring},
ObjectNet \citep{barbu2019objectnet}, and others (e.g.,
\citep{hendrycks2019natural,shankar2019image}). In a similar vein,
\citet{torralba2011unbiased} study model performance when trained on one
standard dataset and tested on another. We omit a detailed discussion of these
works since \sandbox{} is synthetic by nature (and thus less photorealistic than
the aforementioned studies). As shown in Section \ref{sec:realworld-exp}, however,
\sandbox{} is indeed realistic enough to be indicative of real-world performance.

\paragraph{Interpretability, counterfactuals, and model debugging.} 
\sandbox{} can be cast as a method for {\em debugging} vision models that provides users fine-grained control over the rendered scenes and thus enabling them to find
specific modes of failure (cf.  Sections
\ref{sec:evaluation} and \ref{sec:realworld-exp}). Model debugging is also a common goal in
intepretability research, where methods generally seek to provide justification
for model decisions based on either local features (i.e., specific to
the image at hand) or global ones (i.e., general biases of the model). 
Local explanation methods, including saliency maps \citep{simonyan2013deep,
dabkowski2017real, sundararajan2017axiomatic}, surrogate models such as LIME
\citep{ribeiro2016should}, and counterfactual image pairs
\citep{fong2017interpretable,zhu2017object, goyal2019counterfactual}, can
provide insight into specific model decisions but can also be fragile
\citep{ghorbani2019interpretation,melis2018robustness} or misleading with
respect to global model behaviour \cite{adebayo2018sanity,
sundararajan2017axiomatic, adebayo2020debugging, lipton2018mythos}. 
Global interpretability methods include concept-based explanations
\citep{bau2017network, kim2018interpretability, yeh2020completeness,
wong2021leveraging} (though such explanations can often lack causal link to the
features models actually use \citep{goyal2019counterfactual}), but also
encompass many of the robustness studies highlighted earlier in this section,
which can be cast as uncovering global biases of vision models.

\paragraph{Simulated environments and training data.}
Finally, there has been a long line of work on developing simulation platforms
that can serve as both a source of additional (synthetic) training data, and as
a proxy for real-world experimentation. 
Such simulation environments are thus increasingly playing a role in fields 
such as computer vision, robotics, and
reinforcement learning (RL).  
For instance, OpenAI
Gym~\citep{brockman2016openai} and DeepMind Lab \citep{beattie2016deepmind}
provide simulated RL training environments with a fleet of control tasks.
Other frameworks such as UnityML \citep{juliani2020unity} and RoboSuite
\citep{zhu2020robosuite} were subsequently developed to cater to more complex
agent behavior.

In computer vision, the Blender rendering engine \citep{blender} has been 
used
to generate synthetic training data through projects such as BlenderProc
\cite{denninger2019blenderproc} and BlendTorch \cite{heindl2020blendtorch}. 
Similarly, HyperSim \cite{hypersim} is a photorealistic synthetic dataset
focused on multimodal scene understanding.
Another line of work learns optimal simulation
parameters for synthetic data generation according to user-defined objective,
such as minimizing the distribution gap between train and test
environments~\citep{kar2019meta,devaranjan2020meta,behl2020autosimulate}.
Simulators such as AirSim \cite{shah2018airsim}, FlightMare
\cite{song2020flightmare}, and CARLA \cite{dosovitskiy2017carla} 
(built on top of video game engines Unreal Engine and Unity) allow for
collection of synthetic training data for perception and control. 
In robotics, simulators include environments that model  
typical household layouts for robot navigation \cite{kolve2017ai2, 
wu2018building, puig2018virtualhome}, interactive ones where objects that 
can be actuated~\citep{xia2018gibson, xia2020interactive, xiang2020sapien}, 
and those that include support for tasks such as question answering and 
instruction following~\citep{savva2019habitat}.

While some of these platforms may share components with \sandbox{} (e.g., the
physics engine, photorealistic rendering), the do not share the same goals as
\sandbox{}, i.e., diagnosing specific failures in existing models. 

    \section{Conclusion}
    \label{sec:conclusion}
    In this work, we introduced \sandbox, a unified framework 
for diagnosing failure modes in vision models based on high-fidelity 
rendering. 
We demonstrate the utility of \sandbox{} by applying it to a number of model 
debugging use cases---such as understanding classifier sensitivities to 
realistic scene and object perturbations, and discovering model biases.
Further, we show that the debugging analysis done using \sandbox{} in 
simulation is actually predictive of model behavior in the physical world.
Finally, we note that \sandbox{} was designed with extensibility as a priority;
we encourage the community to build upon the framework so as to uncover 
new insights into the vulnerabilities of
vision models.

    \section*{Acknowledgements}
   Work supported in part by the NSF grants CCF-1553428 and CNS-1815221, 
   the Google PhD Fellowship, the Open Philanthropy Project AI Fellowship, 
   and the NDSEG PhD Fellowship.
This material is based upon work supported by the Defense Advanced 
Research
Projects Agency (DARPA) under Contract No. HR001120C0015.

    Research was sponsored by the United States Air Force Research 
    Laboratory
    and the United States Air Force Artificial Intelligence Accelerator and was
    accomplished under Cooperative Agreement Number FA8750-19-2-1000. The views
    and conclusions contained in this document are those of the authors and
    should not be interpreted as representing the official policies, either
    expressed or implied, of the United States Air Force or the U.S. Government.
    The U.S. Government is authorized to reproduce and distribute reprints for
    Government purposes notwithstanding any copyright notation herein. 

    \clearpage
    \printbibliography
    \clearpage
    \appendix
    \section{Implementation and scalability}
\label{app:performance}
In this section, we briefly describe the underlying architecture of \sandbox,
and verify that the system can effectively scale to distributed compute
infrastructure.

\begin{figure*}[!htbp]
    \centering
    \begin{subfigure}{.6\columnwidth}
        \centering
        \includegraphics[width=1\textwidth]{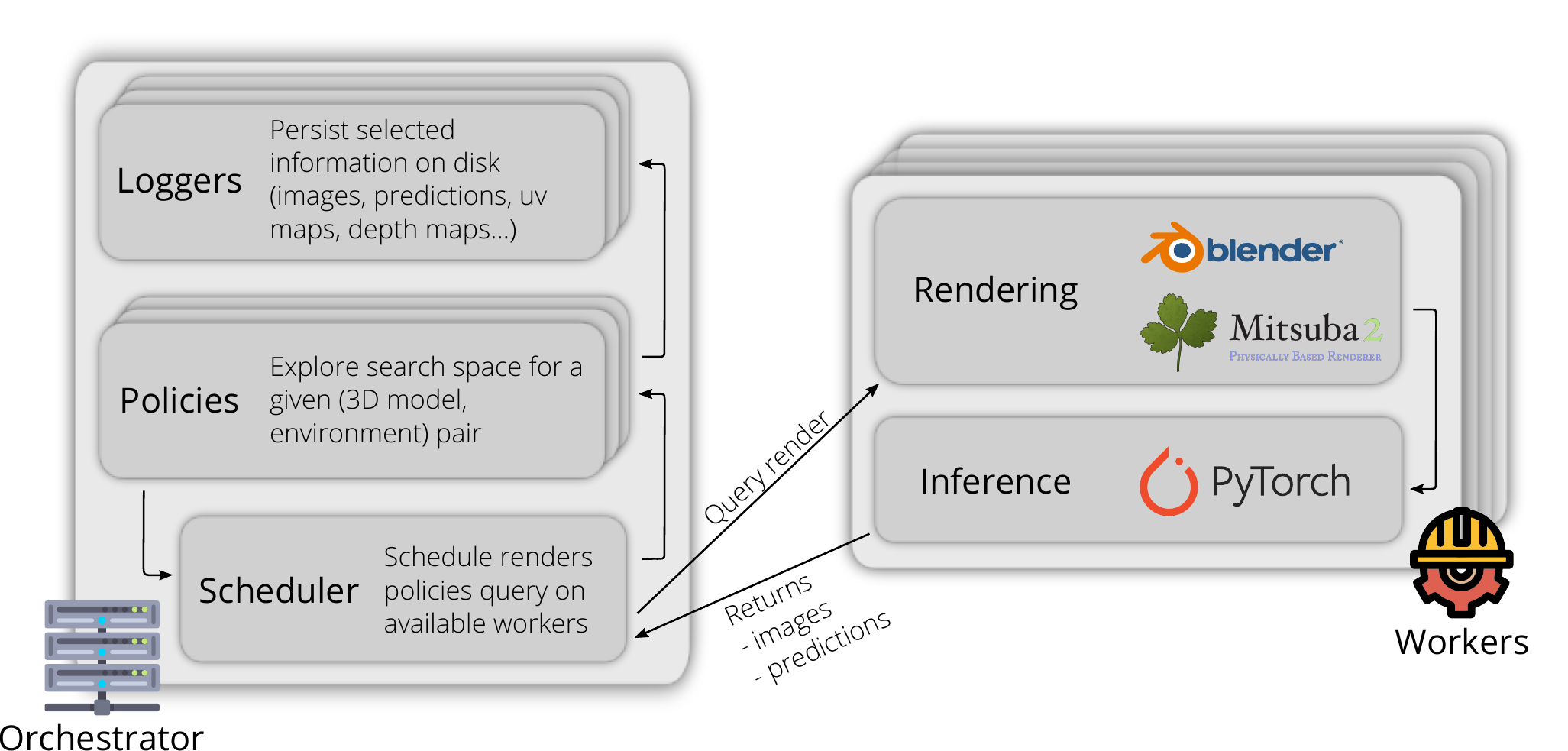}
        \caption{Overview of the architecture of \sandbox}
        \label{fig:architecture}            
    \end{subfigure}
    \begin{subfigure}{.35\columnwidth}
        \includegraphics[width=1\textwidth]{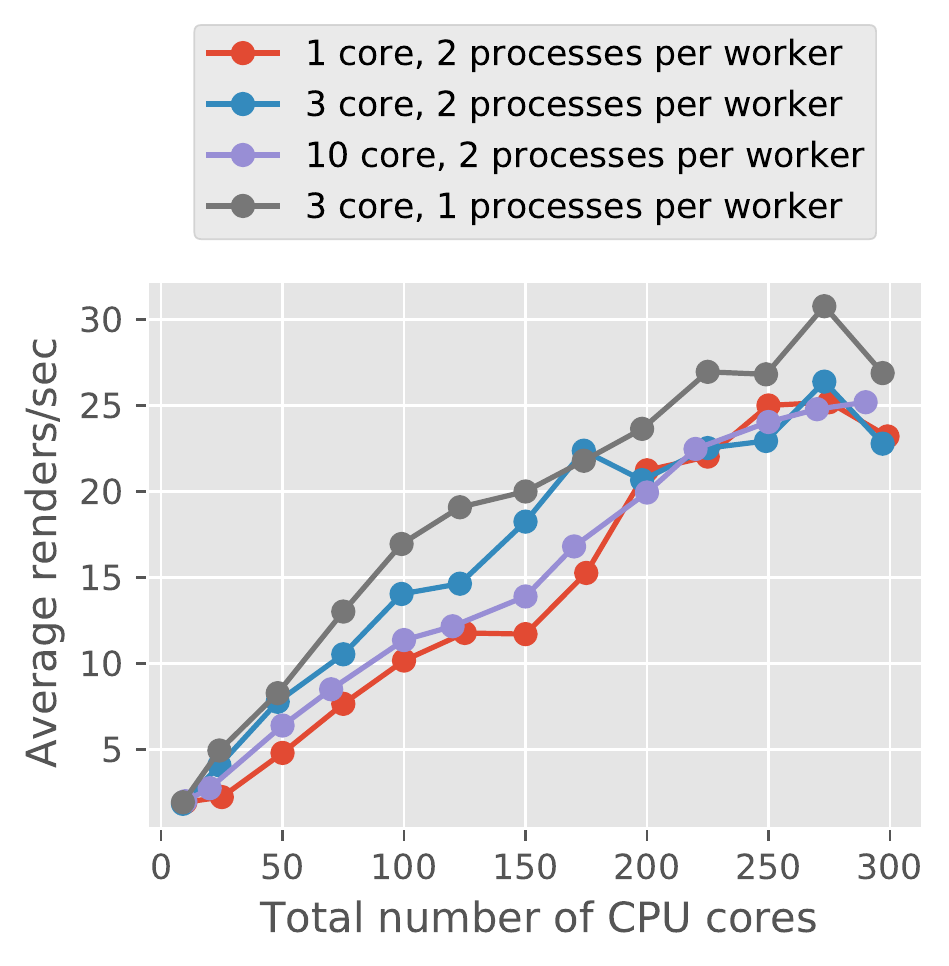}
        \caption{Performance of \sandbox{} on a simple experiment in function
        of the number of CPU cores recruited for renders.}
        \label{fig:perf_exp_speedup}
    \end{subfigure}
\end{figure*}

\paragraph{Architecture.} 
To ensure scalability of this pipeline, we
implement \sandbox{} as a client-server application (cf. Figure
\ref{fig:architecture}). 
The main ``orchestrator'' thread constructs a search space by composing the
user's specified controls, then uses the (user-specified) policy to find the
exact set of 3D configurations that need to be rendered and analyzed.   
It then schedules these configurations across a set of worker nodes, whose 
job is to receive configurations, render them, run inference using the user's
pretrained vision model, and send the results back to the orchestrator node. 
The results are aggregated and written to disk by a logging module. The
dashboard is implemented as a separate entity that reads the log files and
produces a user-friendly web interface for understanding the \sandbox{} results.

\paragraph{Scalability.}
As discussed in Section \ref{sec:design}, in order to perform photo-realistic
rendering at scale, \sandbox{} must be able to leverage many machines (CPU
cores) in parallel. \sandbox{} is designed to allow for this. It can accommodate
as many rendering clients as the user can afford and the rendering efficiency of
\sandbox{} largely scales linearly with available CPU cores (cf.
Figure~\ref{fig:perf_exp_speedup}). Note that the although the user can add as
much rendering clients as they want, the number of actually used clients by the
orchestrator is limited by its number of policy instances. In our paper, we run
a limited number instances of policies (one instance per (env, 3D model) pair)
concurrently to keep the memory of the orchestrator under control. This limits
the scalability of the system as the maximum of renders that has to be done at
any point in time scales with the number of policies of the orchestrator.  Yet,
were able to reach 415 FPS average/800 FPS peak throughput with dummy workers
(no rendering), and around 100 FPS for the main experiments of this paper (e.g.
physical realism experiment) which uses a complex background environment
requiring substantial amount of rendering time (15 secs per image).

\clearpage
\section{Experiment Dashboard}
\label{app:dashboard}

\begin{figure}[!htbp]
    \centering
        \includegraphics[width=\textwidth]{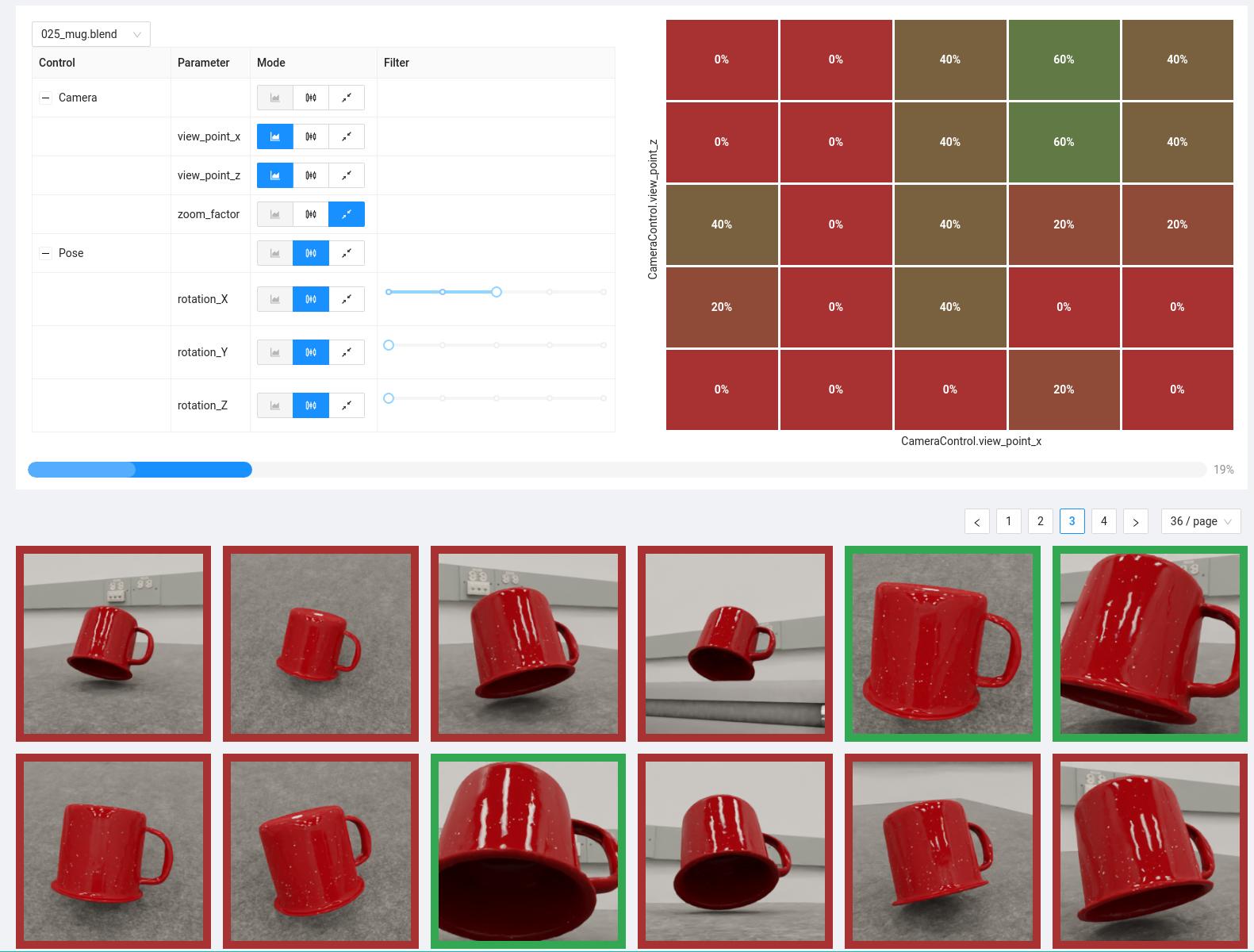}
        \caption{Screenshot of the dashboard used for data exploration.}
       \label{fig:dashboard_screenshot}
\end{figure}

Since experiments usually produce large amounts of data that can be hard to get a sense of,
we created a data visualization dashboard. Given a folder containing the JSON logs of a job,
it offers a user interface to explore the influence of the controls.

For each parameter of each control, we can pick one out three mode:
\begin{itemize}
    \item \textbf{Heat map axis}: This control will be used as the \texttt{x} or \texttt{y}
        axis of the heat map. Exactly two controls should be assigned to this
        mode to enable the visualization. Hovering on cells of the heat map will
        filter all samples falling in that region.
    \item \textbf{Slider}: This mode enables a slider that is used to only select the samples that match exactly this particular value.
    \item \textbf{Aggregate}: do not filter samples based on this parameter
\end{itemize}

\section{iPhone App}
\label{app:iphone_app}
\begin{figure}[h]
    \centering
    \includegraphics[width=0.4\textwidth]{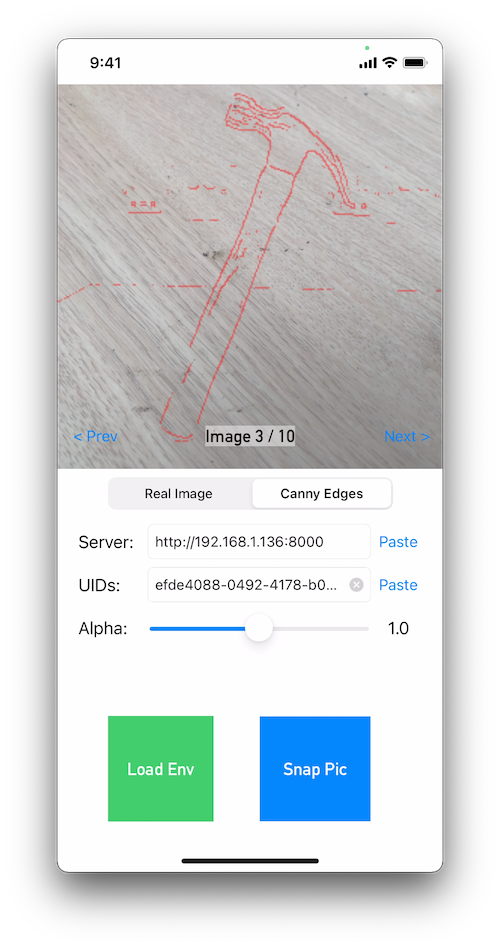}
    \caption{A screenshot of the iOS app used to align objects for the
    physical-world experiment. After starting the dashboard server, the user can
    specify the server location as well as a set of rendering IDs. The
    corresponding renderings will be displayed over a camera view, allowing the
    user to correctly position the object in the frame. The user can adjust the
    object transparency, and can toggle between overlaying the full rendering
    and overlaying just the edges (shown here).}
    \label{fig:ios_app} 
\end{figure}

We developed a native iOS app to help align objects in the physical experiment
(Section \ref{sec:realworld-exp}). The app allows the user to enter one or more
rendering IDs (corresponding to scenes rendered by \sandbox); the app then
brings up a camera with a translucent overlay of either the scene or an
edge-filtered version of the scene (cf. Figure \ref{fig:ios_app}). We used the
app to align the physical object and environment with their intended place in
the rendered scene. The app connects to the same backend serving the experiment
dashboard.

\section{Controls} 
\label{app:controls}

\sandbox{} takes an object-centric perspective, where an object of interest is
spawned on a desired background. The scene mainly consists of the object and a camera. The controls in our pipeline affect this interplay between the scene components through various combinations of properties, which subsequently creates a wide variety of rendered images.  The controls are implemented using the Blender Python API `\texttt{bpy}' that exposes an easy to use framework for controlling Blender. `\texttt{bpy}' primarily exposes a scene context variable, which contains references to the properties of the components such as objects and the camera; thus allowing for easy modification. 

\sandbox{} comes with several predefined controls that are ready to use (see \url{https://3db.github.io/3db/}). Nevertheless, users are able (and encouraged) to implement custom controls for their use-cases.

\clearpage
\section{Additional Experiments Details}

We refer the reader to our package \url{https://github.com/3db/3db} for all source code, 3D models, HDRIs, and 
config files used in the experiments of this paper.

For all experiments we used the pre-trained ImageNet ResNet-18 included in
\texttt{torchvision}. In this section we will describe, for each experiment
the specific 3D-models and environments used by \sandbox{}
to generate the results.

\begin{figure}[!htbp]
    \centering
    \begin{subfigure}[b]{.45\columnwidth}
        \centering
        \includegraphics[width=\textwidth]{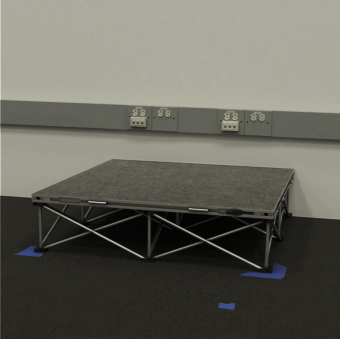}
        \caption{Synthetic}
        \label{fig:msr_studio_synth}
    \end{subfigure}
    \hfill
    \begin{subfigure}[b]{.45\columnwidth}
        \centering
        \includegraphics[width=\textwidth]{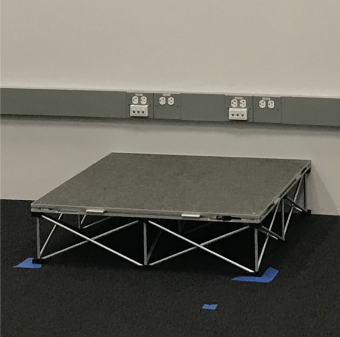}
        \caption{Real picture (iPhone 12 Pro)}
        \label{fig:msr_studio_real}
    \end{subfigure}
       \caption{Studio used for the real-world experiments (Section
       \ref{sec:realworld-exp}).}
       \label{fig:msr_studio}
\end{figure}

\label{app:exp_setup}
\subsection{Sensitivity to image backgrounds (Section \ref{sec:backgrounds_sensitivity})}

\subsubsection{Analysing a subset of backgrdounds}
\label{app:subset-backgrounds}
\paragraph{Models:} We collected 19 3D-models in total. On top of the models shown on
figure \ref{fig:texture_swap_histograms}, we used models for: (1) an orange, (2)
two different toy power drills, (3) a baseball ball, (4) a tennis ball, (5) a
golf ball, (6) a running shoe, (7) a sandal and (8) a toy gun. Some of these models are from \texttt{YCB} \cite{calli2015benchmarking} and the rest are purchased from \texttt{amazon.com} and then put through a 3D scanner to get corresponding meshes.

\paragraph{Environments:} We sourced 20 2k HDRI from the website
\url{https://hdrihaven.com}. In particular we used: \texttt{abandoned\_workshop,
adams\_place\_bridge, altanka, aristea\_wreck, \\bush\_restaurant, cabin,
derelict\_overpass, dusseldorf\_bridge, factory\_yard, gray\_pier,
greenwich\_park\_03, kiara\_7\_late-afternoon, kloppenheim\_06, rathaus,
roofless\_ruins, secluded\_beach, small\_hangar\_02, stadium\_01, studio\_small\_02,
studio\_small\_04}.

\subsubsection{Analyzing all backgrounds with the “coffee mug” model.}
\paragraph{Models:} We used a single model: the coffee mug, in order to keep computational resources under control.

\paragraph{Environments:} We used 408 HDRIs
from \url{https://hdrihaven.com/} with a 2K resolution.

\subsection{Texture-shape bias (section \ref{sec:texture-exp})}

\begin{figure}
\centering
\includegraphics[width=0.8\textwidth]{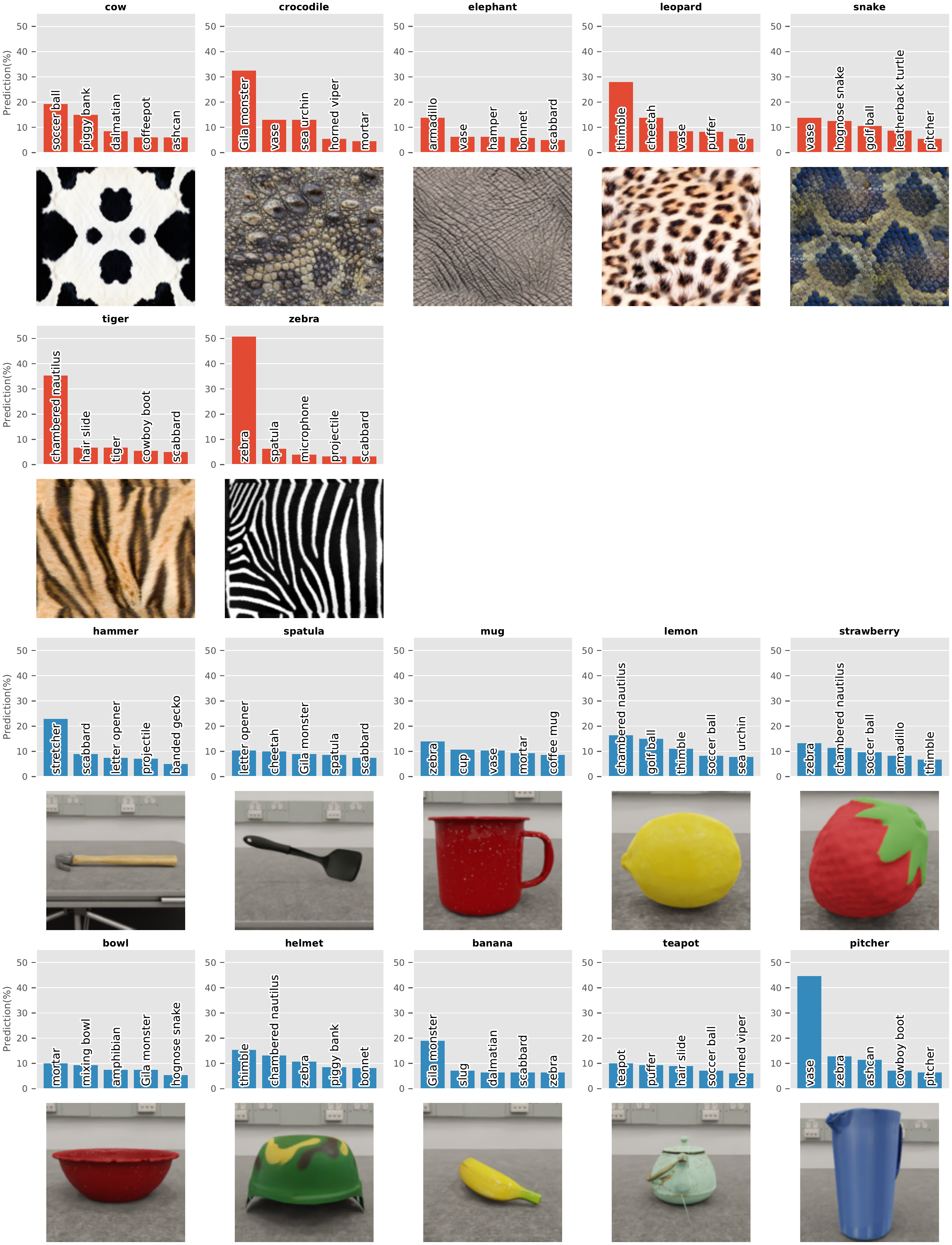}
\caption{Additional examples of the experiment in Figure~\ref{fig:texture_swap_histograms_short}. Distribution of classifier predictions after the texture of the 3D object model 
is altered. In the top rows, we visualize the most frequently predicted classes for each texture (averaged over all objects). In the bottom rows, we visualize the most
frequently predicted classes for each object (averaged over all textures). We find that the model tends to predict based on the texture more often than based
on the object.}
\label{fig:texture_swap_histograms}
\end{figure}

\paragraph{Textures:} To replace the original materials, we collected 7 textures
on the internet and we modified them to make them seamlessly tilable. These textures are shown on Figure \ref{fig:texture_swap_histograms}.

\paragraph{Models:} We used all models that are shown on Figure
\ref{fig:texture_swap_histograms}.

\paragraph{Environments:} We used the virtual studio
environment (Figure \ref{fig:msr_studio}).

\subsection{Orientation and scale dependence (Section \ref{sec:orientation_scale_dependence})}
We use the same models and environments that are used in Appendix~\ref{app:subset-backgrounds}.

\subsection{3D models Heatmaps (Figure \ref{fig:heatmap_visualizations})}

\paragraph{Models:} For this experiment we used the set of models shown on
Figure \ref{fig:texture_swap_histograms}.

\paragraph{Environments:} We used the virtual studio environment (see Figure
\ref{fig:msr_studio}).

\subsection{Case study: using 3DB to dive deeper (Section \ref{subsec:mug_fluid})}
\label{subsec:exp_setup_mug_fluid}

\paragraph{Models:} We only used the mug since this
experiment is mug specific.

\paragraph{Environments:} We used the sudio set shown on Figure
\ref{fig:msr_studio}.

\subsection{Physical realism (Section \ref{sec:realworld-exp})}
\label{subsec:exp_setup_real_world}

\paragraph{Real-world pictures:} All images were taken with an handheld Apple
iPhone 12 Pro. To help us align the shots we used the application described in
appendix \ref{app:iphone_app}.

\paragraph{Models:} We used the models shown in Figure
\ref{fig:real_life_exp_samples}.

\paragraph{Environments:} The environment shown on Figure \ref{fig:msr_studio}
was especially designed for this experiment. The goal was to have an environment
that matches our studio as closely as possible. The geometry and materials were
carefully reproduced using reference pictures. The lighting was reproduce
through a high resolution HDRI map.

\subsection{Performance scaling (Appendix \ref{app:performance})}
\label{subsec:exp_setup_perf_scaling}

The only relevant details for this experiment are the fact that we ran 10
policies (at most 5 concurrently). Each policy consisted of 1000 renders using a
2k HDRI as environment.

\clearpage
\section{Omitted Figures}
\label{app:omitted_figures}
\begin{figure}[!htbp]
    \centering
    \includegraphics[width=.3\columnwidth]{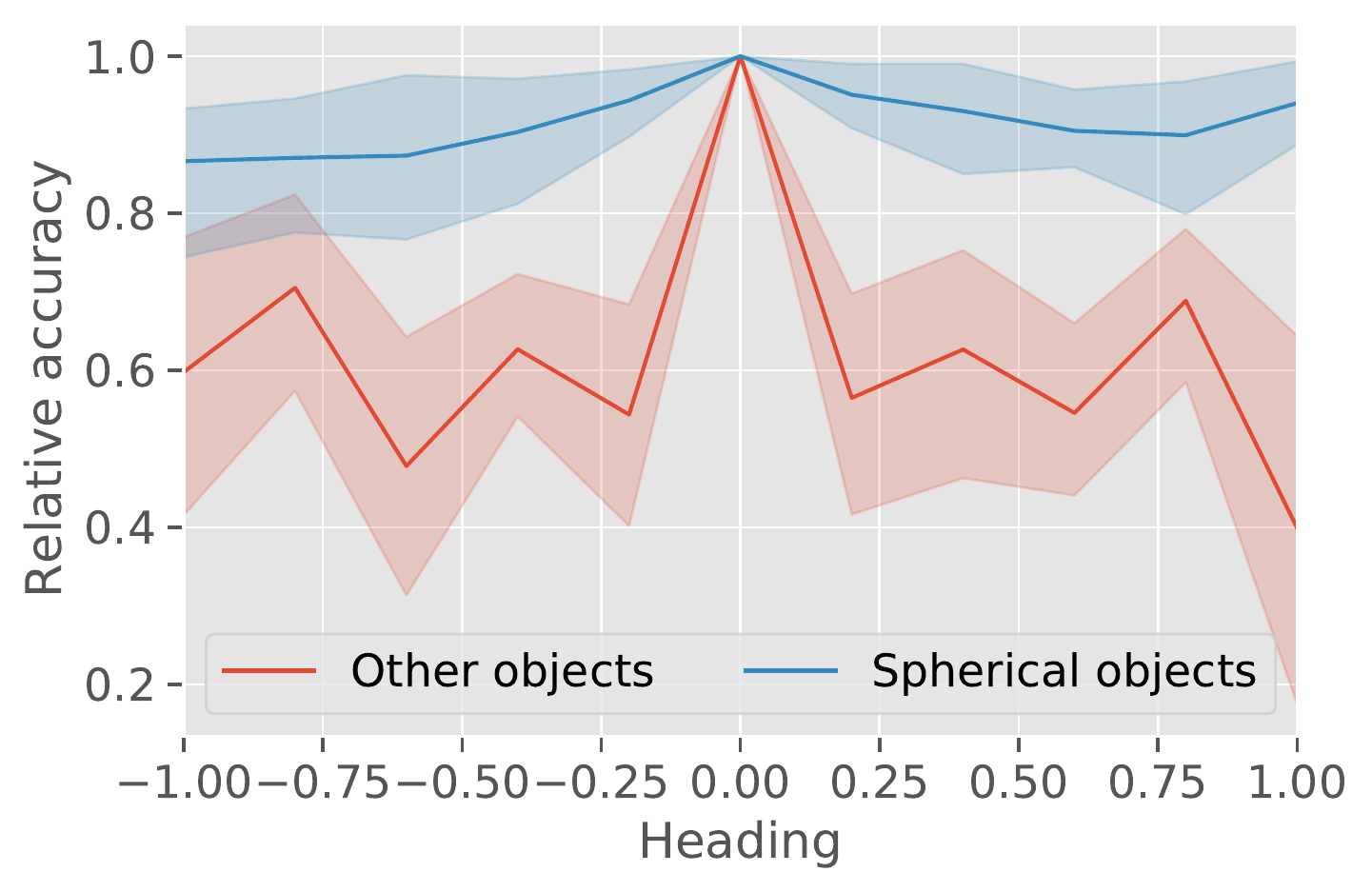}
    \includegraphics[width=.3\columnwidth]{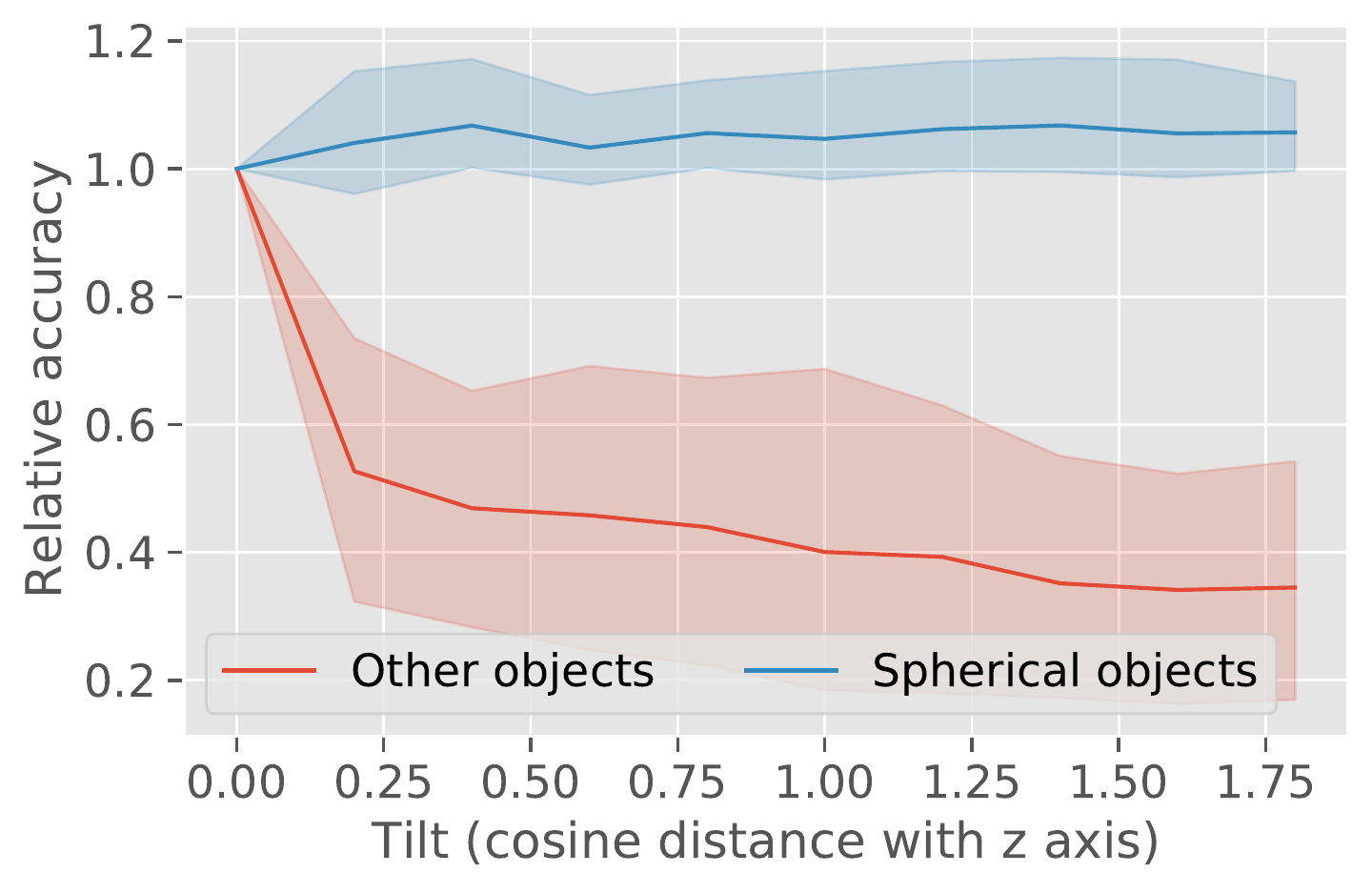}
    \includegraphics[width=.3\columnwidth]{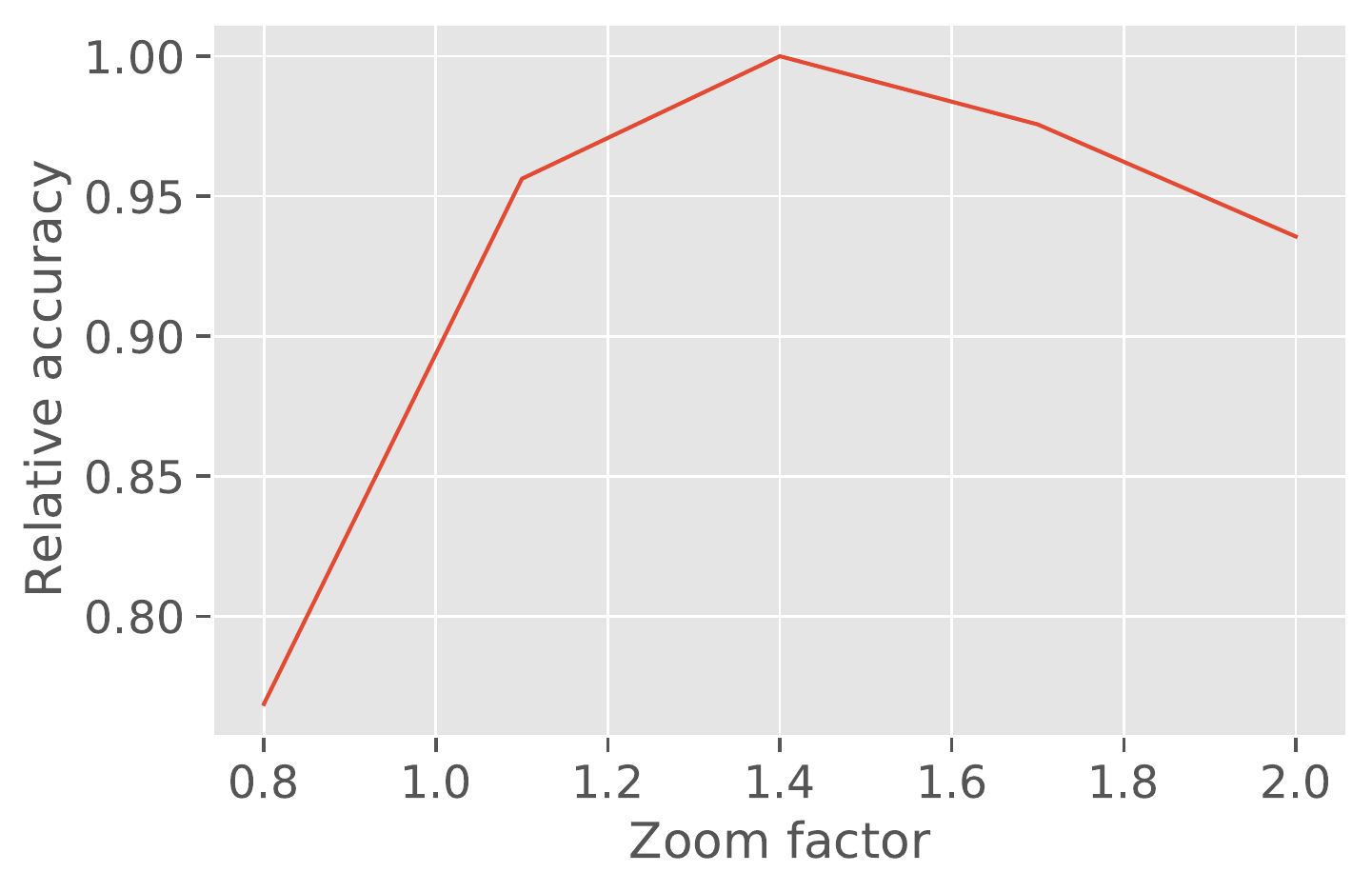}
    \caption{ Additonal plots to Figure~\ref{fig:imgnet_orientation_dep}. We plot the distribution of model accuracy as a function of object heading 
    (\emph{top}), tilt (\emph{middle}) and zoom (\emph{bottom}), aggregated 
    over variations in controls. For heading and tilt, we separately evaluate accuracy for (non-)spherical objects. Notice how the performance of the model degrades for non-spherical objects as the heading/tilt changes, but not for spherical objects. Also notice how the performance depends on the zoom level of the camera (how large the object is in the frame).}
\end{figure}
\begin{figure}[!htbp]
\centering
\begin{subfigure}{.4\columnwidth}
    \includegraphics[width=1\textwidth]{./figures/mug_liquid_experiment_samples.png}
    \caption{Sample of the images rendered for the experiment presented in section
    \ref{subsec:mug_fluid}.}
    \label{fig:mug_liquid_experiment_samples_2}        
\end{subfigure}
\hspace{1cm}
\begin{subfigure}{.5\columnwidth}
    \centering
    \includegraphics[width=\textwidth]{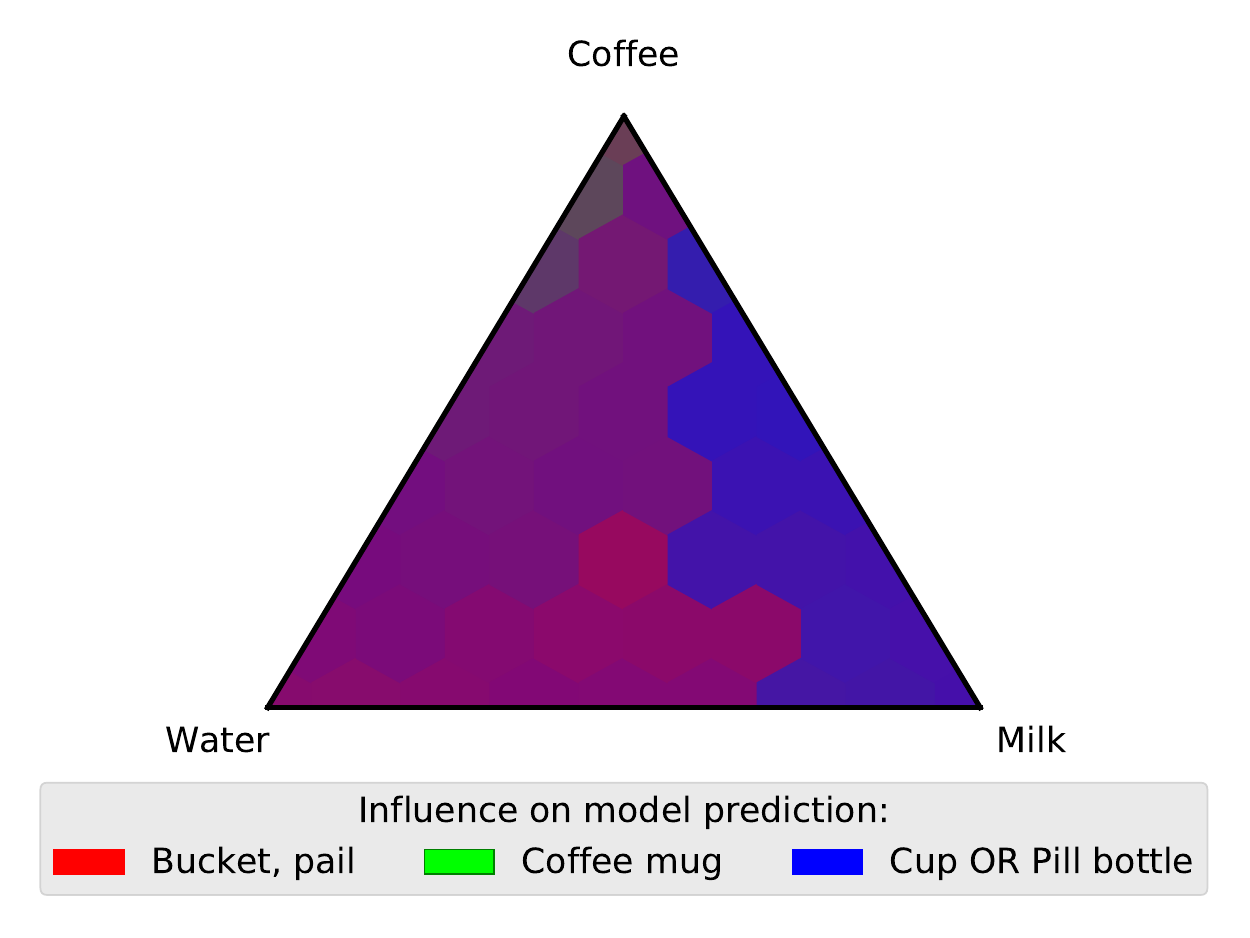}
    \caption{\label{fig:mug_liquid_experiment_simplex_raw} Un-normalized version of
    \autoref{fig:mug_fluid_results}-(b).}
\end{subfigure}
\caption{Additional illustration for the mug liquid experiment of Figure~\ref{fig:mug_fluid_results}. This figure shows the
correlation of the liquid mixture in the mug on the prediction of the model, averaged over random viewpoints }
\end{figure}

    \end{document}